\documentclass[journal,10pt,draftclsnofoot,onecolumn]{IEEEtran}
\usepackage{graphics,multirow,amsmath,amssymb,textcomp,subfigure,multirow,xspace,arydshln,cite}
\usepackage[]{graphicx}

\usepackage{enumerate}
\usepackage{amsmath, amsfonts}
\usepackage{layout}

\usepackage[ruled,boxed,linesnumbered]{algorithm2e}
\usepackage[pagebackref=true,breaklinks=true,letterpaper=true,colorlinks,bookmarks=false,linkcolor=black,citecolor=black,urlcolor=blue]{hyperref}

\usepackage{stfloats}
\usepackage{url}
\usepackage{multirow}
\usepackage{graphicx}
\usepackage{afterpage}
\usepackage{pbox}
\usepackage{enumitem}

\usepackage{amsfonts}

\usepackage{algpseudocode}
\LinesNumberedHidden
\DontPrintSemicolon

\makeatletter

\DeclareRobustCommand\onedot{\futurelet\@let@token\@onedot}
\def\@onedot{\ifx\@let@token.\else.\null\fi\xspace}

\makeatother

\begin{document}

\def\x{{\mathbf x}}
\def\v{{\mathbf v}}
\def\q{{\mathbf q}}
\def\xh{{\hat{\x}}}
\def\qh{{\hat{\q}}}
\def\xi{\x_i}

\def\oneb{\mathrm{1}}
\def\one{\underbar{$ \oneb $}}
\def\y{{\mathbf y}}
\def\z{{\mathbf z}}
\def\p{{\mathbf p}}
\def\q{{\mathbf q}}
\def\U{{\mathbf U}}

\def\Gam{{\mathbf \Gamma}}
\def\Q{{\mathbf Q}}
\def\D{{\mathbf D}}
\def\W{{\mathbf W}}
\def\Z{{\mathbf Z}}
\def\Zh{{\mathbf \hat \Z }}
\def\V{{\mathbf V}}
\def\S{{\mathbf S}}
\def\G{{\mathbf G}}
\def\Dh{{\hat{\D}}}
\def\H{\mathbf {H}}
 
\def\L{{\mathrm L}}
\def\I{{\mathrm I}}
\def\b{{\mathbf b}}
\def\W{{\mathbf W}}
\def\A{{\mathbf A}}
\def\B{{\mathrm B}}
\def\Ai{\A_i}
\def\T{{\mathrm T}}
\def\Ti{\T_i}
\def\R{{\mathbf R}}
\def\P{{\mathbf P}}
\def\r{{\mathrm r}}
\def\c{{\mathrm c}}
\def\hhsigma{{\hat{\sigma}}}
\def\RR{{\mathbb R}}
\def\h{{\mathbf h}}

\title{RAISR: Rapid and Accurate Image \\ Super Resolution}
\author{Yaniv~Romano, John~Isidoro,~and~Peyman Milanfar,~\IEEEmembership{Fellow,~IEEE}
\thanks{More results can be found in the supplementary material: \url{https://goo.gl/D0ETxG}. This work was performed while the first author was an intern at Google Research. Y. Romano is with the Department of Electrical Engineering, Technion -- Israel Institute of Technology, Technion City, Haifa 32000, Israel. E-mail address: yromano@tx.technion.ac.il. J. Isidoro and P. Milanfar are with Google Research, 1600 Amphitheatre Pkwy, Mountain View, CA 94043. E-mail addresses: \{isidoro,milanfar\}@google.com.}}

\maketitle
\begin{abstract}

Given an image, we wish to produce an image of larger size with significantly more pixels and higher image quality. This is generally known as the Single Image Super-Resolution (SISR) problem. The idea is that with sufficient training data (corresponding pairs of low and high resolution images) we can learn set of filters (i.e. a mapping) that when applied to given image that is not in the training set, will produce a higher resolution version of it, where the learning is preferably low complexity. In our proposed approach, the run-time is more than one to two orders of magnitude faster than the best competing methods currently available, while producing results comparable or better than state-of-the-art.

A closely related topic is image sharpening and contrast enhancement, i.e., improving the visual quality of a blurry image by amplifying the underlying details (a wide range of frequencies). Our approach additionally includes an extremely efficient way to produce an image that is significantly sharper than the input blurry one, without introducing artifacts such as halos and noise amplification. We illustrate how this effective sharpening algorithm, in addition to being of independent interest, can be used as a pre-processing step to induce the learning of more effective upscaling filters with built-in sharpening and contrast enhancement effect. 

\end{abstract}



\section{Introduction}
\label{sec:background}

Single Image Super Resolution (SISR) is the process of estimating a High-Resolution (HR) version of a Low-Resolution (LR) input image. This is a well-studied problem, which comes up in practice in many applications, such as zoom-in of still and text images, conversion of LR images/videos to high definition screens, and more. The linear degradation model of the SISR problem is formulated by
\begin{align}
\label{eq:degradation}
\z = \D_{s}\H\x,
\end{align}
where $ \z\in\RR^{M\times N} $ is the input image, $ \x\in\RR^{Ms\times Ns} $ is the unknown HR image, both are held in lexicographic ordering. The linear operator $ \H \in \RR^{MNs^2\times MNs^2}$ blurs the image $ \x $, followed by a decimation in a factor of $ s $ in each axis, which is the outcome of the multiplication by $ \D_{s} \in \RR^{MN\times MN s^2}$. In the SR task, the goal is to recover the unknown underlying image $ \x $ from the known measurement $ \z $. Note that, in real world scenarios, the degradation model can be non-linear (e.g. due to compression) or even unknown, and may also include noise. 
 

The basic methods for upscaling a single image are the linear interpolators, including the nearest-neighbor, bilinear and bicubic  \cite{LIN_INT_REF1,LIN_INT_REF2}. These methods are widely used due to their simplicity and low complexity, as the interpolation kernels (upscaling filters) are not adaptive to the content of the image. However, naturally, these linear methods are limited in reconstructing complex structures, often times result in pronounced aliasing artifacts and over-smoothed regions. In the last decade powerful image priors were developed, e.g., the self-similarity \cite{buades_nonlocal,hiro_kernel,bm3d,hiro_tip2009}, sparsity \cite{aharon_ksvd,YangSR,zeyde2012single,ANR,timofte2014a+,romano2014single}, and Gaussian Mixtures \cite{zoran2011learning}, leading to high quality restoration with the cost of increased complexity. 

In this paper we concentrate on example-based methods \cite{YangSR,zeyde2012single,yang2012coupled,peleg2014statistical,timofte2014a+,dong2014learning,dong2015,dai2015jointly}, which have drawn a lot of attention in recent years. The core idea behind these methods is to utilize an external database of images and learn a mapping from LR patches to their HR counterparts. In the learning stage, LR-HR pairs of image patches are synthetically generated, e.g., for $ 2\times $ upscaling, a typical size of the HR patch is $ 6\times 6 $ and the one of the synthetically downscaled LR patch is $ 3\times 3 $. Then, the desired mapping is learned and regularized using various local image priors.

The sparsity model is one such prior \cite{YangSR,zeyde2012single}, where the learning mechanism results in a compact (sparse) representation of pairs of LR and HR patches over learned dictionaries. Put differently, per each LR patch, these methods construct a non-linear adaptive filter (formulated as a projection matrix), which is a combination of a few basis elements (the learned dictionary atoms) that best fit to the input patch. Applying the filter that is tailored to the LR patch leads to the desired upscaling effect.

The Anchored Neighborhood Regression (ANR) \cite{ANR} keeps the high quality reconstruction of \cite{YangSR} and 
\cite{zeyde2012single} while achieving a significant speed-up in runtime. This is done by replacing the sparse-coding step that computes the compact representation of each patch over the learned dictionaries, with set of pre-computed projection matrices (filters), which are the outcome of ridge regression problems. 
As such, at runtime, instead of applying sparse-coding, ANR suggest searching for the nearest atom to the LR patch, followed by a multiplication by the corresponding pre-computed projection matrix.  A follow-up work, called A+ \cite{timofte2014a+}, improves the performance of ANR by learning regressors not only from the nearest dictionary atoms, but also from the locally nearest training samples, leading to state-of-the-art restoration. 

SRCNN \cite{dong2014learning} is another efficient example-based approach that builds upon deep Convolutional Neural Network (CNN) \cite{lecun1989backpropagation}, and learns an end-to-end mapping from LR images to their HR counterparts. Note that, differently from sparsity-based techniques, SRCNN does not explicitly learn the dictionaries for modeling the patches. In this case, the model is implicitly learned by the hidden convolutional layers.  

The above mentioned SISR methods result in impressive restoration, but with the cost of (relatively) high computational complexity. In this paper we suggest a learning-based framework, called RAISR, which produces high quality restoration while being two orders of magnitude faster than the current leading algorithms, with extremely low memory requirements. 

The core idea behind RAISR is to enhance the quality of a very cheap (e.g. bilinear) interpolation method by applying a set of pre-learned filters on the image patches, chosen by an efficient hashing mechanism. Note that the filters are learned based on pairs of LR and HR training image patches, and the hashing is done by estimating the local gradients' statistics. As a final step, in order to avoid artifacts, the initial upscaled image and its filtered version are locally blended by applying a weighted average, where the weights are a function of a structure descriptor. We harness the Census Transform (CT) \cite{zabih1994non} for the blending task, as it is extremely fast and cheap descriptor of the image structure which can be utilized to detect structure deformations that occur due to the filtering step.

A closely related topic to SISR is image sharpening, aiming to amplify the structures/details of a blurry image. The basic sharpening techniques apply a linear filter on the image, as in the case of unsharp masking \cite{polesel2000image} or Difference of Gaussians (DoG) \cite{marr1980theory,winnemoller2012xdog}. These techniques are highly effective in terms of complexity, but tend to introduce artifacts such as over-sharpening, gradient reversals, noise amplification, and more. Similarly to SISR, improved results can be obtained by relying on patch priors, where the sensitivity to the content/structure of the image is the key for artifact-free enhancement \cite{zhang2008adaptive,he2010guided,talebi2014nonlocal,kheradmand2015nonlinear,liu2015joint}. For example, with the cost of increased complexity compared to the linear approach, the edge-aware bilateral filter \cite{tomasi,durand2002fast}, Non-Local Means \cite{buades_nonlocal} and guided filter \cite{he2010guided} produce impressive sharpening effect. 

As a way to generate high-quality sharp images, one can learn a mapping from LR images to their \emph{sharpened} HR versions, thus achieving a built-in sharpening/contrast-enhancement effect ''for free''. Furthermore, the learning stage is not limited to a linear degradation model (as in Eq. (\ref{eq:degradation})), as such, learning a mapping from \emph{compressed} LR images to their \emph{sharpened} HR versions can be easily done, leading to an "all in one" mechanism that not only increases the image resolution, but also reduces compression artifacts and enhances the contrast of the image.

Triggered by this observation, we develop a sharpener as well, which is of independent interest. The proposed sharpener is highly efficient and able to enhance both fine details (high frequencies) and the overall contrast of the image (mid-low frequencies). The proposed method has almost similar complexity to the linear sharpeners, while being competitive with far more complex techniques. The suggested sharpener is based on applying DoG filters \cite{marr1980theory, winnemoller2012xdog} on the image, which are capable to enhance a wide range of frequencies. Next, a CT-based structure-aware blending step is applied as a way to prevent artifacts due to the added content-aware property (similar mechanism to the one suggested in the context of SISR). 

This paper is organized as follows: In Section \ref{sec:global} we describe the global learning and upscaling scheme, formulating the core engine of RAISR. In Section \ref{sec:refinement} we refine the global approach by integrating the initial upscaling kernel to the learning scheme. In Section \ref{sec:hash} we describe the overall learning and upscaling framework, including the hashing and blending steps. The sharpening algorithm is detailed in Section \ref{sec:sharpener}. Experiments are brought in Section \ref{sec:experiments}, comparing the proposed upscaling and sharpening algorithm with state-of-the-art methods. Conclusions and future research directions are given in Section \ref{sec:conclusion}.

\begin{figure}
	\begin{center}
		\subfigure[Learning Stage]{\includegraphics[scale=0.45]{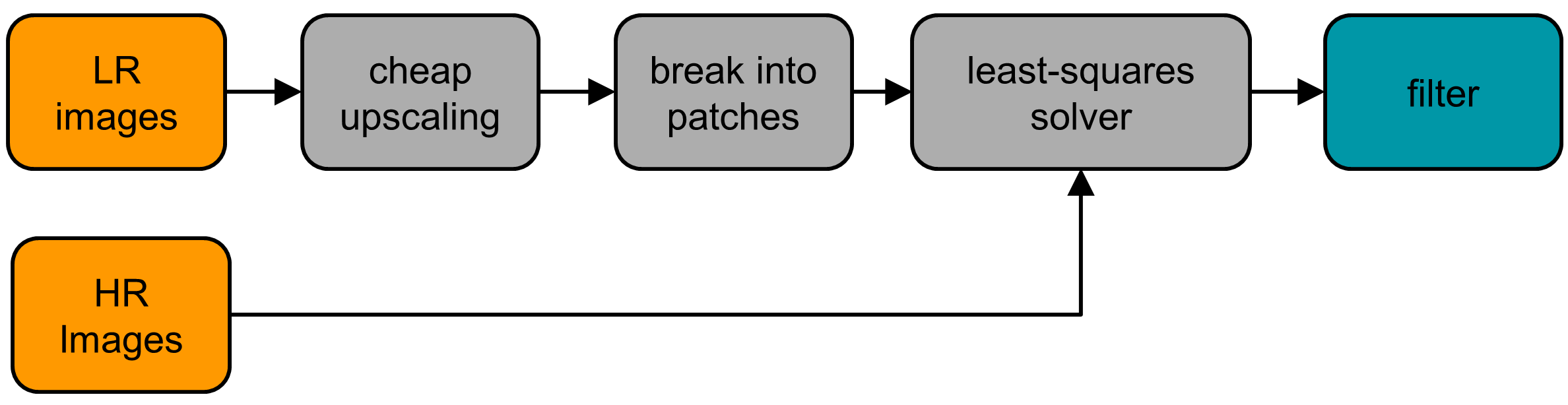}} \\
		\subfigure[Upscaling Stage]{\includegraphics[scale=0.45]{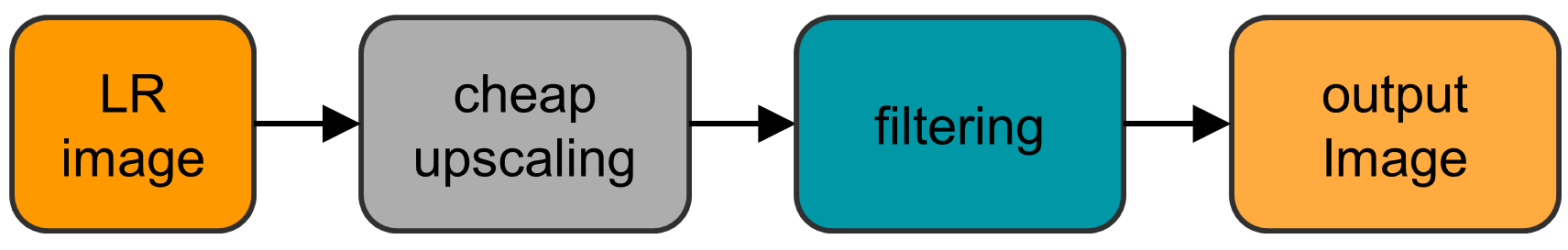}}
	\end{center}
	\vspace{-0.2in}
	\caption{The basic learning and application scheme of a global filter that maps LR images to their HR versions.}
	\label{fig:globalblock}
\end{figure}

\section{First Steps: Global Filter Learning}
\label{sec:global}

Given an initial (e.g. bilinear in our case) upscaled versions of the training database images, \mbox{$\y_i \in \RR^{M\times N}$}, with $i = 1, \cdots, L$, we aim to learn a $d\times d$ filter $h$ that minimizes the Euclidean distance between  the collection $\{ \y_i\}$ and the desired training HR images $\{\x_i\}$. Formally, this is done by solving a least-squares minimization problem
\begin{equation}
\min_{\h} \sum_{i=1}^{L} \| \A_i \h - \b_i \|_2^2
\label{eq:globalLS}
\end{equation}
\noindent where $\h \in \RR^{d^2} $ denotes the filter $h \in \RR^{d \times d}$ in vector-notation; $\A_i \in \RR^{MN\times d^2}$ is a matrix, composed of patches of size $d\times d$, extracted from the image $\y_i$, each patch forming a row in the matrix. The vector $\b_i \in \RR^{MN} $ is composed of pixels from $\x_i$, corresponding to the center coordinates of $\y_i$ patches. The block diagram, demonstrating the core idea of the learning process is given in Fig. \ref{fig:globalblock}a.

In practice, the matrix $\A$ can be very large, so we employ two separate approaches to control the computational complexity of estimating the filter. First, in general not all available patches needs to be used in order to obtain a reliable estimate. In fact, we typically construct $\A_i$ and $\b_i$ by sampling $K$ patches/pixels from the images on a fixed grid, where $K \ll MN$. Second, the minimization of the least-squares problem, formulated in Eq. \eqref{eq:globalLS}, can be recast in a way that significantly reduces both memory and computational requirements. To simplify the exposition, the following discussion is given in the context of filter learning based on just one image, but extending the idea to several images and filters is trivial. The proposed approach results in an efficient solution for the learning stage where the memory requirements are only on the order of the size of the learned filter. The solution is based on the observation that instead of minimizing Eq. \eqref{eq:globalLS}, we can minimize
\begin{equation}
\min_{\h}  \| \Q \h -\V \|_2^2,
\label{eq:compactLS}
\end{equation}
where $ \Q = \A^T\A $ and $ \V = \A^T\b $.

Notice that $\Q $ is a small $d^2\times d^2$ matrix, thus requiring relatively little memory. The same observation is valid for $\V$ that requires less memory than holding the vector $\b$. Furthermore, based on the inherent definition of matrix-matrix and matrix-vector multiplications, we in fact avoid holding the whole matrix (and vector) in memory. More specifically, $\Q $ can be computed cumulatively by summing chunks of rows (for example sub matrices $\A_j \in \RR^{q \times d^2}$, $ q \ll MN $), which can be multiplied independently, followed by an accumulation step; i.e.
\begin{equation}
 \Q = \A^T\A  =  \sum_{j} \A_j^T\A_j 
\label{eq:accumulateAtA}
\end{equation}
\noindent The same observation is true for matrix-vector multiplication
\begin{equation}
 \V = \A^T\b  =  \sum_{j} \A_j^T\b_j, 
\label{eq:accumulateAtb}
\end{equation}
where $ \b_j \in \RR^{q} $ is a portion of the vector $ \b $, corresponding to the matrix $ \A_j $. Thus, the complexity of the proposed learning scheme in terms of memory is very low -- it is in the order of the filter size. Moreover, using this observation we can parallelize the computation of $ \A_j^T\A_j  $ and $ \A_j^T\b_j $, leading to a speedup in the runtime. As for the least squares solver itself, minimizing Eq. \eqref{eq:compactLS} can be done efficiently since $\Q $ is a positive semi-definite matrix, which perfectly suits a fast conjugate gradients solver \cite{barrett1994templates}.

To summarize, the learning stage is efficient both in terms of the memory requirements and ability to parallelize. As displayed in \mbox{Fig. \ref{fig:globalblock}b}, at run-time, given a LR image (that is not in the training set), we produce its HR approximation by first interpolating it using the same cheap upscaling method (e.g. bilinear) that is used in the learning stage, followed by a filtering step with the pre-learned filter.

\section{Refining the Cheap Upscaling Kernel: Dealing with Aliasing}
\label{sec:refinement}

The ''cheap'' upscaling method we employ as a first step, can be any method, including a non-linear one. However, in order to keep the low complexity of the proposed approach,  we use the bilinear interpolator as the initial upscaling method\footnote{We also restrict the discussion mainly to the case of $2\times$ upscaling to keep the discussion straightforward. Extensions will be discussed at the end of this section.}. Inspired by the work in \cite{peleg2014statistical}, whatever the choice of the initial upscaling method, we make the observation that when aliasing is present the input LR image, the output of the initial upscaler will generally not be shift-invariant to this aliasing.

\begin{figure}
	\begin{center}
		\includegraphics[width=4in]{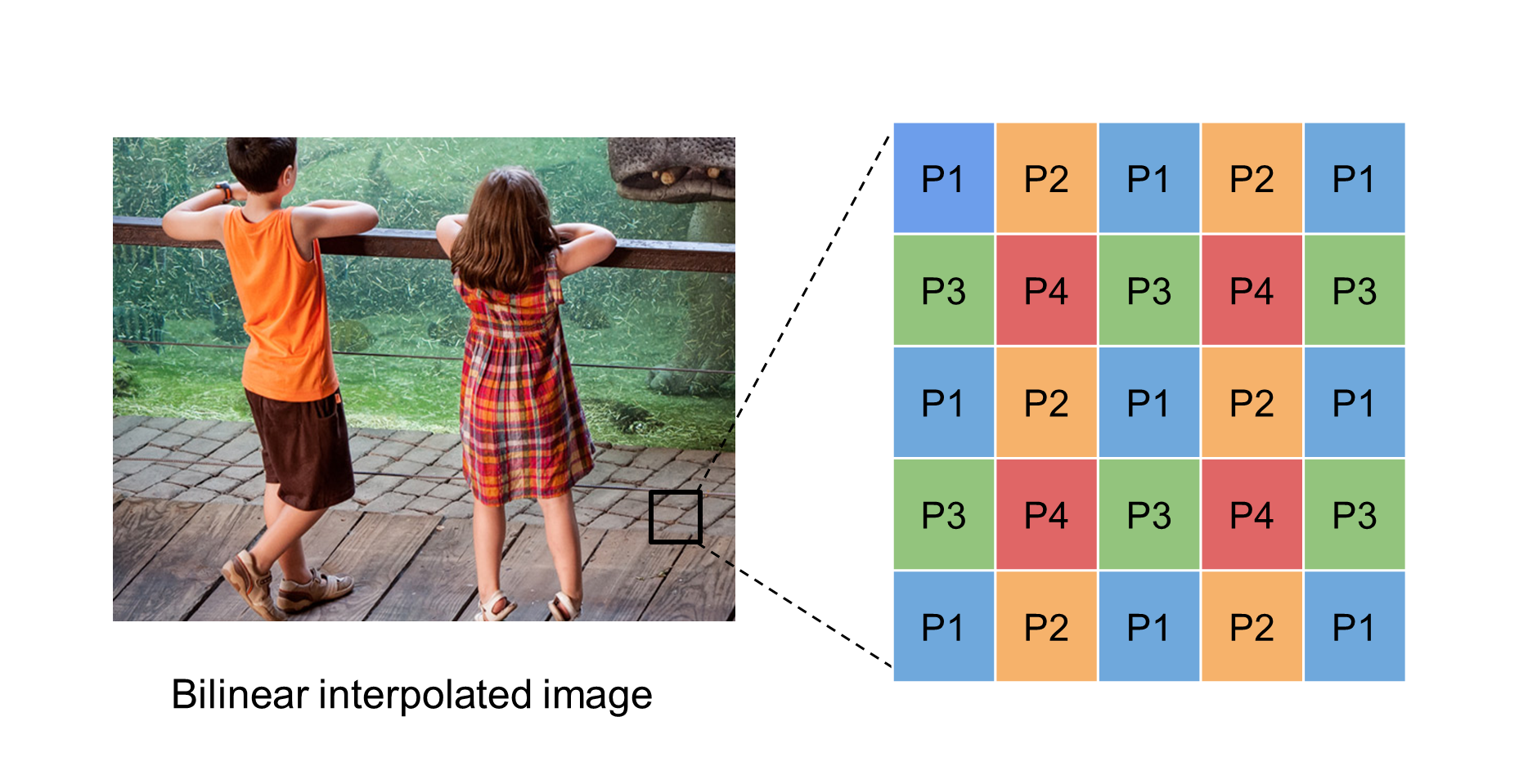}
	\end{center}
	\vspace{-0.2in}
	\caption{Bilinear upscaling by a factor of $2$ in each axis. There are four types of pixels, denoted by P1-P4, corresponding to the four kernels that are applied during the bilinear interpolation.}
	\label{fig:fourfilters}
\end{figure}

\begin{figure}
	\begin{center}
		\includegraphics[width=4.5in]{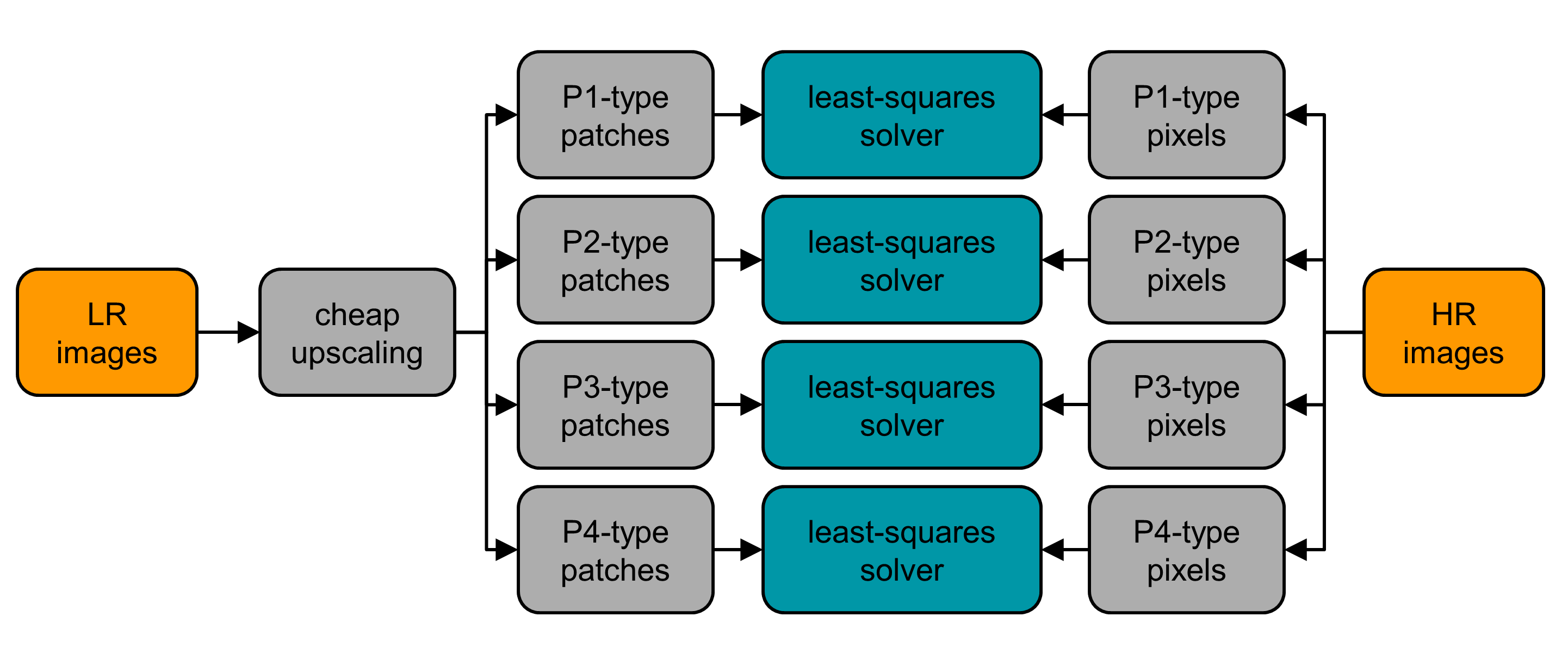}
	\end{center}
	\vspace{-0.3in}
	\caption{Spatially varying learning scheme of four global filters, taking into consideration the internal structure of the bilinear.}
	\label{fig:variantLearning}
\end{figure}

\begin{figure}
	\begin{center}
		\subfigure[P1-Filter]{\includegraphics[width=1.3in, height=1.3in]{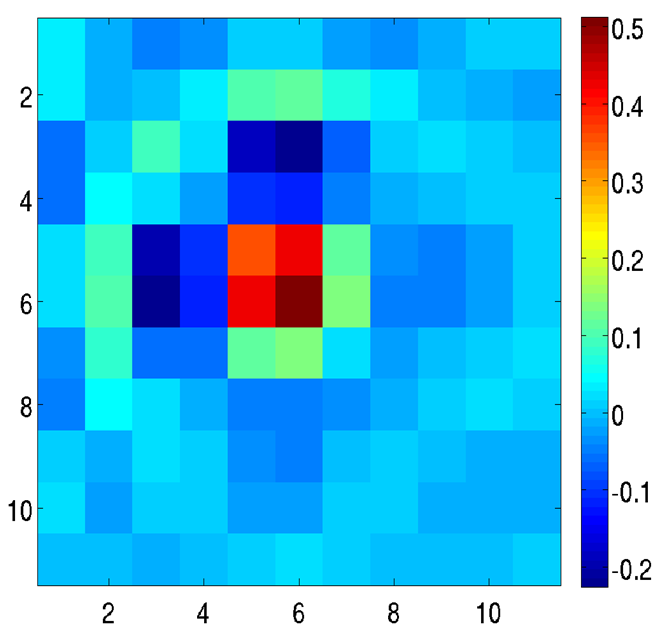}}
		\hfill
		\subfigure[P2-Filter]{\includegraphics[width=1.3in, height=1.3in]{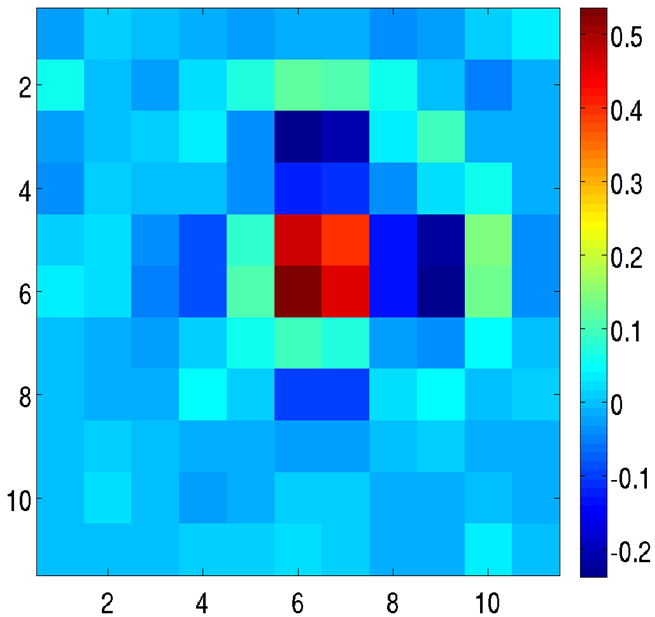}}
		\hfill
		\subfigure[P3-Filter]{\includegraphics[width=1.3in, height=1.3in]{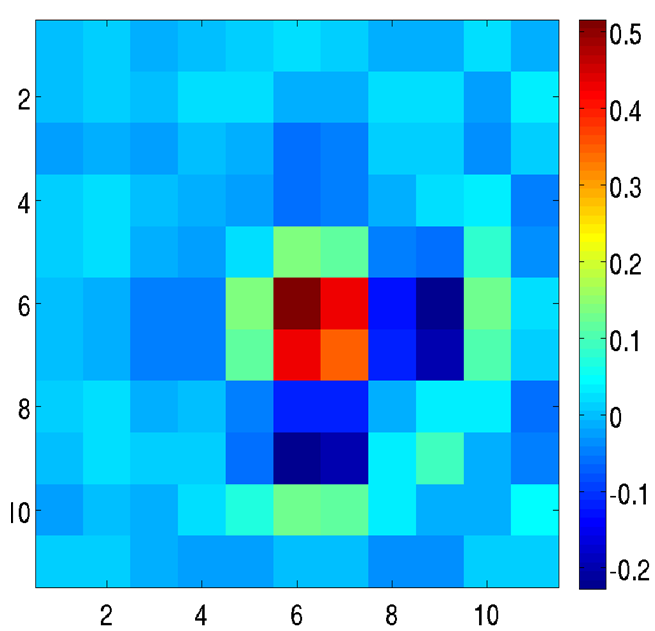}}
		\hfill
		\subfigure[P4-Filter]{\includegraphics[width=1.3in, height=1.3in]{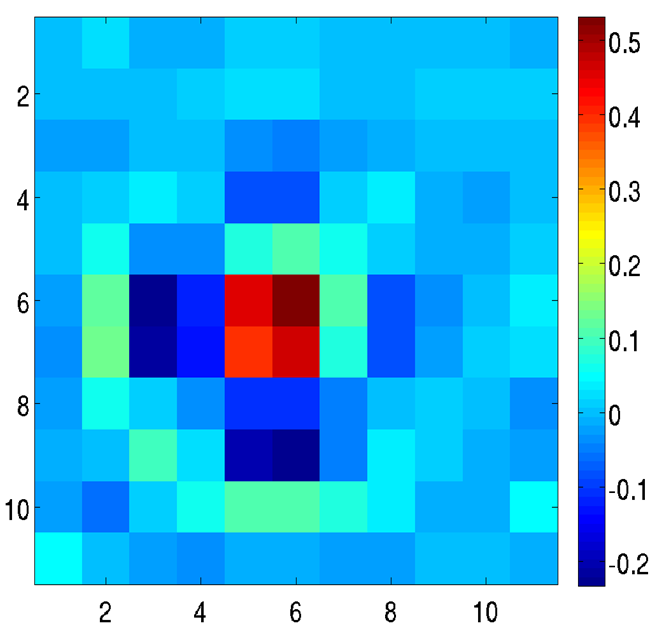}} \\
		
		\subfigure[P1-Magnitude spectrum]{\includegraphics[width=1.3in, height=1.3in]{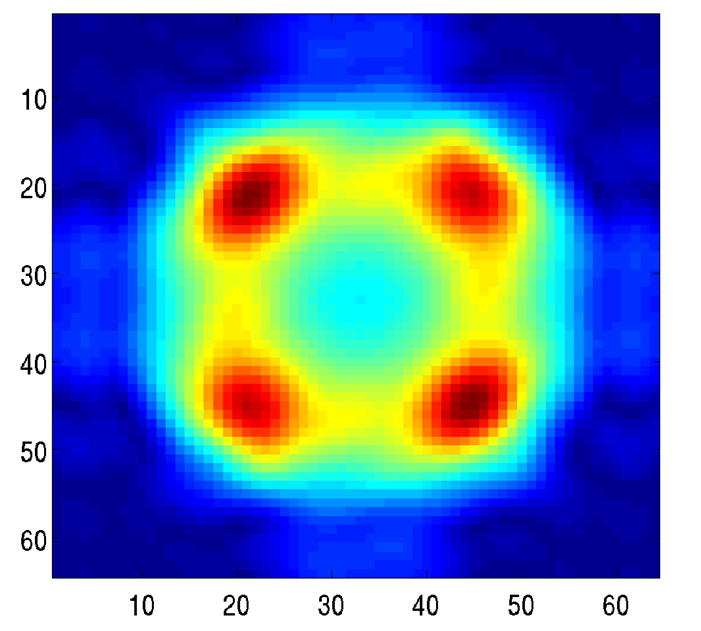}}
		\hfill
		\subfigure[P2-Magnitude spectrum]{\includegraphics[width=1.3in, height=1.3in]{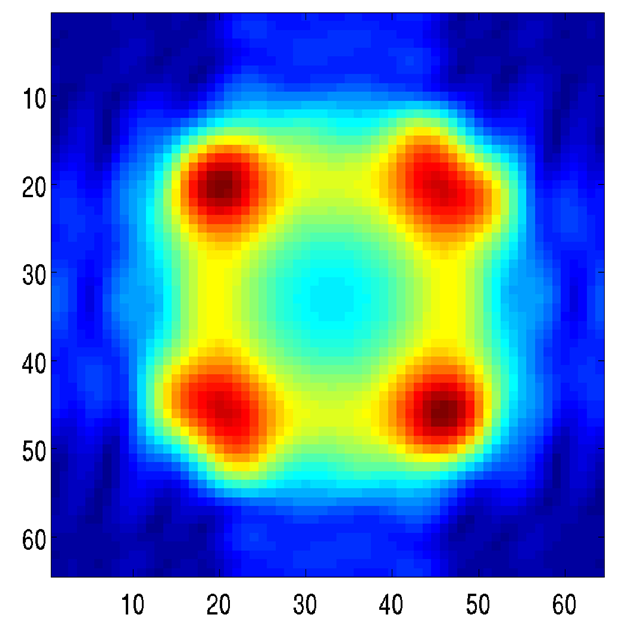}}
		\hfill
		\subfigure[P3-Magnitude spectrum]{\includegraphics[width=1.3in, height=1.3in]{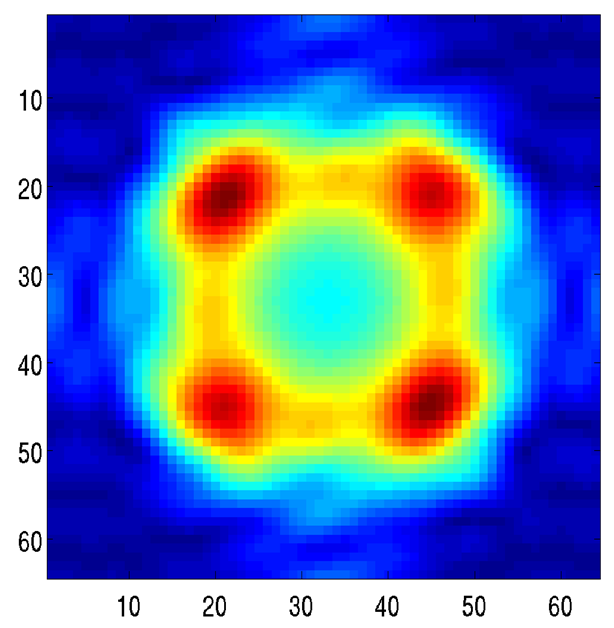}}
		\hfill
		\subfigure[P4-Magnitude spectrum]{\includegraphics[width=1.3in, height=1.3in]{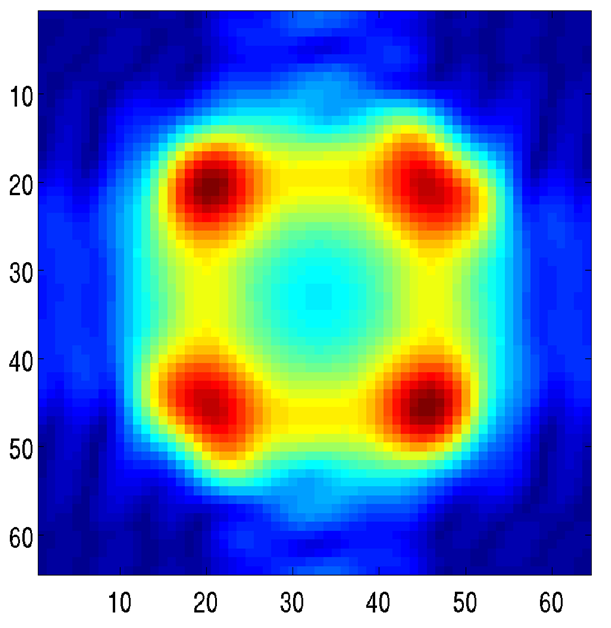}}
	\end{center}
	\vspace{-0.1in}
	\caption{ Visualization of the four global filters, corresponding to P1-P4 type of pixels, in the pixel domain (a-d), along with their magnitude in the frequency domain (e-f), where the warmer the color, the larger the value. The filters are learned on Fig. \ref{fig:fourfilters} image. }
	\label{fig:visFilters}
\end{figure}

As illustrated in Fig. \ref{fig:fourfilters}, in the case of upscaling by a factor of 2 in each axis, the interpolation weights of the bilinear kernel vary according to the pixel's location. As can be seen, there are four possible kernels that are applied on the LR image according to the type of the pixel, denoted by P1-P4. Since a convolution of two linear filters can be unified into one filter (in our case, the first is the bilinear and the second is the pre-learned one)\footnote{This observation is a promising way to further speed up the algorithm and reduce the overall complexity.}, we should learn four different filters, corresponding to the four possible types of pixels, as demonstrated in Fig. \ref{fig:variantLearning}.

The importance of this observation is illustrated in \mbox{Fig. \ref{fig:visFilters}}, which plots examples of actual learned filters, along with their magnitude in the frequency domain. The obtained filters act like bandpass filters, amplifying the mid-frequencies, and suppressing the high-frequencies (which contain aliasing components) of the interpolated image. The learned filters have similar magnitude response (Fig. \ref{fig:visFilters}e-\ref{fig:visFilters}h), but different phase response (Fig. \ref{fig:visFilters}a-\ref{fig:visFilters}d), standing in agreement with the four different shifted versions of the interpolation kernels.

\begin{figure}
	\begin{center}
		\includegraphics[width=4.5in]{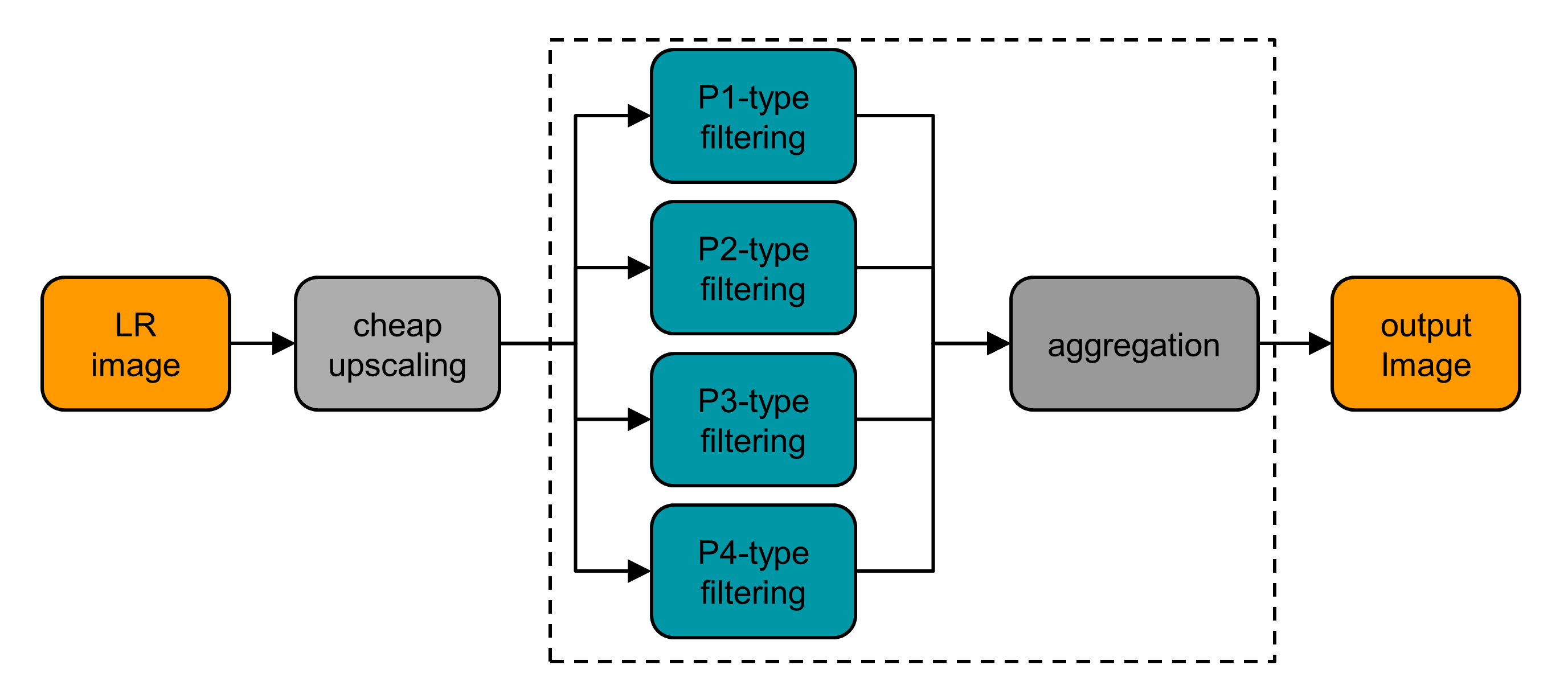}
	\end{center}
	\vspace{-0.3in}
	\caption{Applying the four spatially varying pre-learned filters on a LR image.}
	\label{fig:variantUpscaling}
\end{figure}

On the application side, similarly to the core/naive upscaling idea, we first upscale the LR image using the bilinear interpolator. Then, differently from the naive approach, we apply the pre-learned filters according to the type of the pixel, followed by an aggregation step that simply combines the outcome of the filtered patches (resulting in a pixel) to an image. This process is illustrated in Fig. \ref{fig:variantUpscaling}.

Notice that a similar observation holds for upscaling by any other integer factor $ s $. For example, upscaling by a factor of 3 implies that we should learn 9 filters, one per each pixel-type. Similarly, when upscaling by a factor of 4, there are 16 types of pixels. As already mentioned, in order to keep the flow of the explanations, we will concentrate on the $ 2\times $ scenario since the generalization to other scaling factors is straightforward.

\section{RAISR: Hashing-Based Learning and Upscaling}
\label{sec:hash}

Generally speaking, the global image filtering is fast and cheap, as it implies the application of one filter per patch. Since the learning scheme reduces the Euclidean distance between the HR and the interpolated version of the LR images, the global filtering has the ability to improve the restoration performance of various linear upscaling methods. However, the global approach described so far is weaker than the state-of-the-art algorithms, e.g., sparsity-based methods \cite{YangSR,zeyde2012single,ANR,timofte2014a+} or the neural networks based ones \cite{dong2014learning} that build upon large amount of parameters, minimizing highly nonlinear cost functions. In contrast to these methods, the global approach is not adaptive to the content of the image, and its learning stage estimates only a small amount of parameters.

\begin{figure}
	\begin{center}
				\subfigure[Learning Stage]{\includegraphics[scale=0.5]{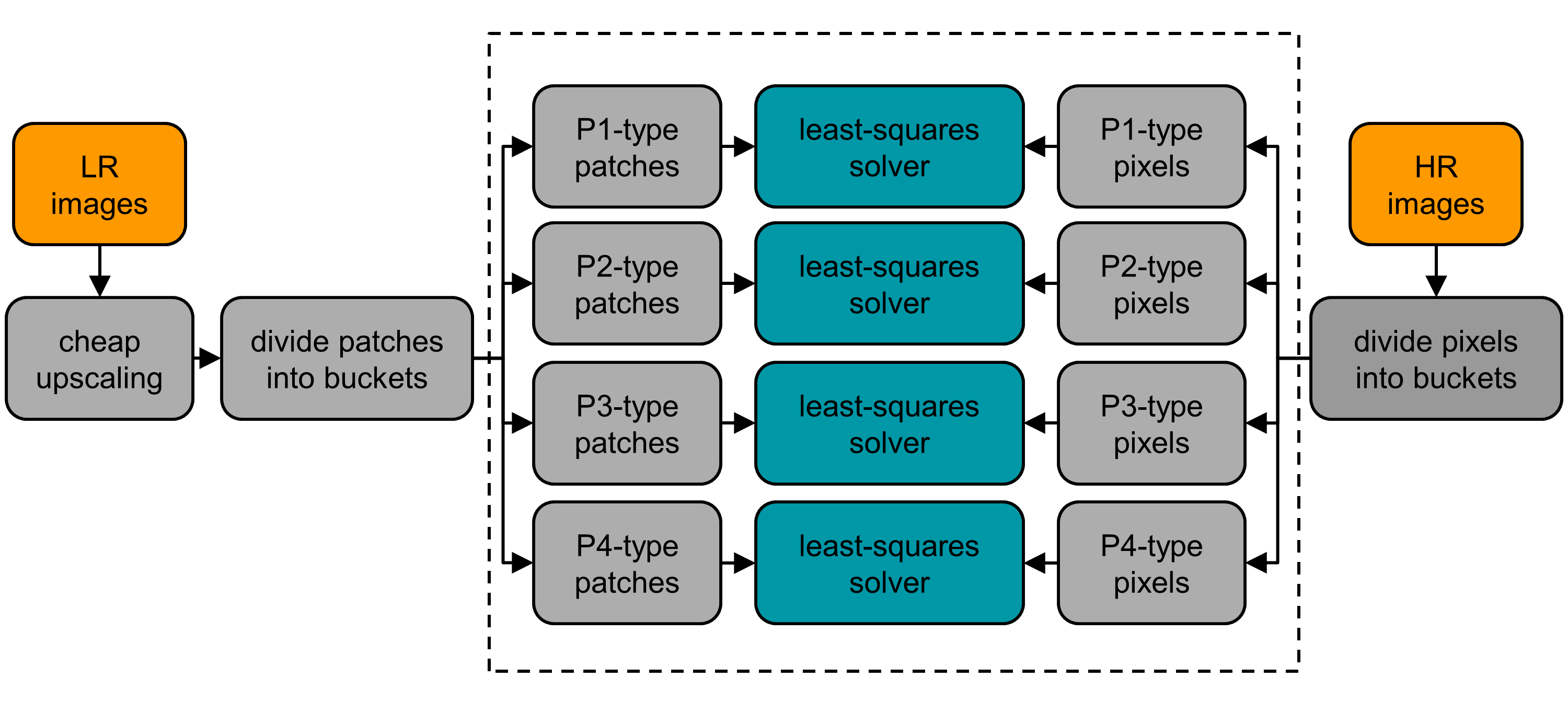}}
				\subfigure[Upscaling Stage]{\includegraphics[scale=0.5]{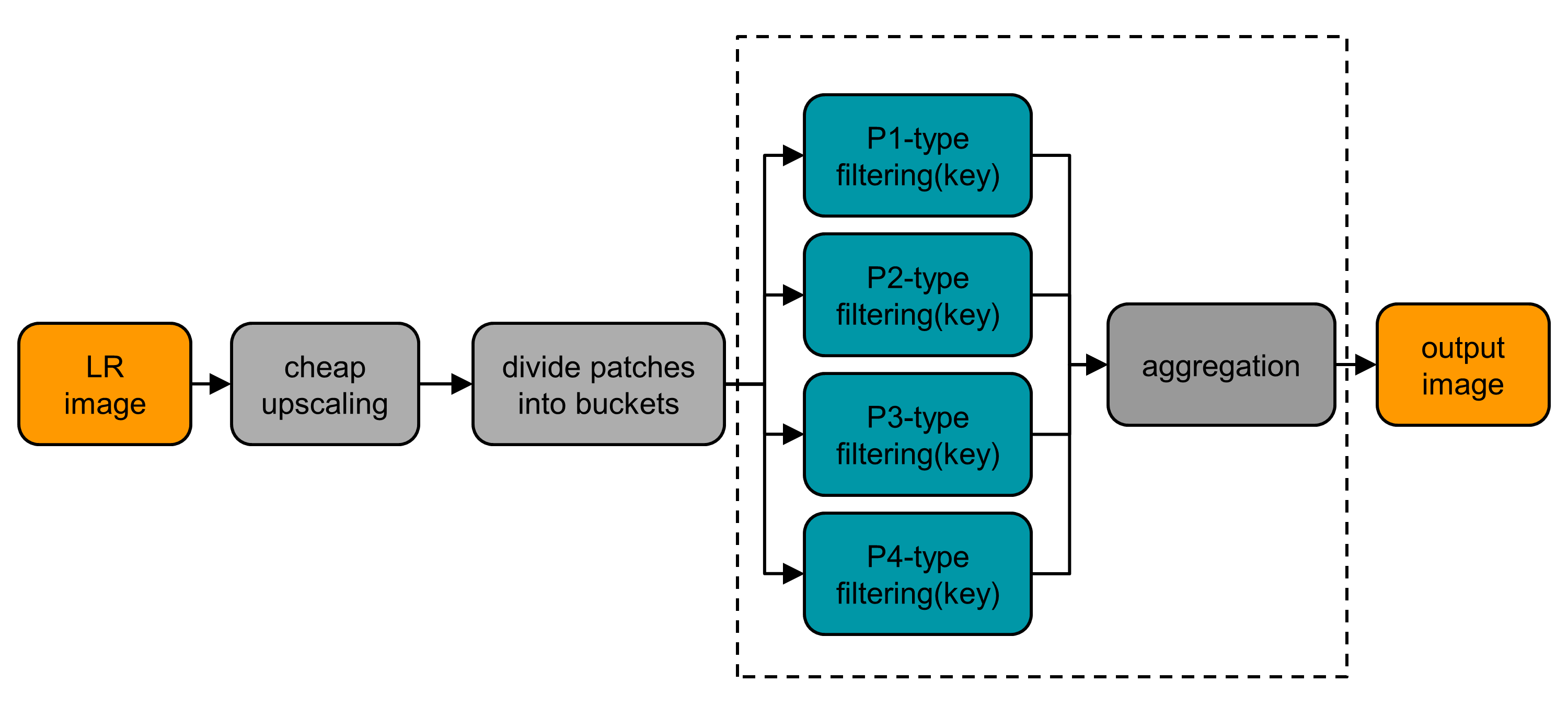}}

	\end{center}
	\caption{Hashing based learning and upscaling schemes. We suggest dividing the patches into ''buckets'', where each bucket contains patches with similar geometry (can be considered as a cheap clustering method). Then, a least squares fitting is applied per each bucket and possible shift. At run-time the hash-table key is computed per each patch, leading to the corresponding pre-learned locally adaptive filters.}
	\label{fig:bucketLearningUpscaling}
\end{figure}
%

Adaptivity to the image content can be achieved by dividing the image patches into clusters, and constructing an appropriate filter per each cluster (e.g. as done in \cite{ANR,timofte2014a+}). However, the clustering implies the increase of the overall complexity of the algorithm, which is an outgrowth that we want to avoid. Therefore, instead of applying ''expensive'' clustering (e.g. K-means\cite{jeong2010training}, GMM \cite{yu2012solving,papyan2016multi}, dictionary learning \cite{YangSR,zeyde2012single,yang2012coupled,peleg2014statistical,timofte2014a+,dong2014learning,dong2015,dai2015jointly}), we suggest using an efficient hashing approach, leading to adaptive filtering that keeps the low complexity of the linear filtering. More specifically, the local adaptivity is achieved by dividing the image patches into groups (called ''buckets'') based on an informative and ''cheap'' geometry measures, which utilize the statistics of the gradients (a detailed description is given in Section \ref{sec:angle}). Then, similarly to the global approach, we also learn four filters, \emph{but this time per each bucket}. As a consequence, the proposed learning scheme results in a hash-table of filters, where the hash-table keys are a function of the local gradients, and the hash-table entries are the corresponding pre-learned filters. An overview of the proposed hashing-based learning is shown in \mbox{Fig. \ref{fig:bucketLearningUpscaling}a}.

Given the hash-table, containing filters per quantized edge-statistic descriptor (more details in Section \ref{sec:angle}), the upscaling procedure becomes very effective. Following Fig. \ref{fig:bucketLearningUpscaling}b, we compute the hash-table key per each patch of the initial interpolated image, pointing to the relevant filters (four filters, one per patch-type), to be applied on the corresponding patch.

Similarly to the global learning process (see Section \ref{sec:global}), we utilize the matrix-matrix and matrix-vector multiplications once again. Per each bucket $ q $, we learn a filter $ \h_q $ by minimizing the following cost function
\begin{equation}
\min_{\h_q}  \| \A_q^T\A_q \h_q -\A_q^T\b_q \|_2^2,
\label{eq:compactLS_bucket}
\end{equation}
where $ \A_q $ and $ \b_q $ are the patches and pixels that belong to the $ q $-th bucket. In this case, the low memory requirements of the proposed learning process are crucial, especially for large hash-table that requires millions of examples to produce a reliable estimate for the filters. As a consequence, by utilizing the observation described in Section \ref{sec:global}, we perform a sub-matrix accumulation on a sub-image block basis, leading to a learning process that can handle any desired number of examples.

\subsection{Hash-Table Keys: Local Gradient Statistics (Angle, Strength, Coherence)}
\label{sec:angle}

Naturally, there are many possible local geometry measures that can be used as the hash-table keys, whereas the statistics of the gradients has a major influence on the proposed approach. 
We suggest evaluating the local gradient characteristics via eigenanalysis \cite{feng}, which yields the gradient's angle and information about the strength and coherence of the nearby gradients. Eigenanalysis also helps in cases of thin lines, stripes and other scenarios that the mean gradient might be zero, yet the neighborhood exhibits a strong directionality.  

The direction, strength and coherence are computed by utilizing the $ \sqrt{n}\times \sqrt{n} $ surroundings of each pixel, i.e., for the $ k $-th pixel we consider all the pixels that are located at $ k_1,...,k_n $.
The basic approach starts with a computation of $ 2\times n $ matrix, composed from the horizontal and vertical gradients, $ g_{x} $ and $ g_{y} $, of the surroundings of the $ k $-th pixel, expressed by
\begin{equation}
\G_k = \begin{bmatrix}
g_{x_{k_1}} & g_{y_{k_1}} \\
\vdots & \vdots  \\
g_{x_{k_n}} & g_{y_{k_n}} \end{bmatrix}.
\end{equation}
As stated in \cite{feng}, the local gradient statistics can be computed using the Singular Value Decomposition (SVD) of this matrix. The right  vector corresponds to the gradient orientation, and the two singular values indicate the strength and spread of the gradients. Since the work is being performed per-pixel, we hereby focus on efficiency. We can compute those characteristics more efficiently using an eigen-decomposition of $ \G_k^T\G_k $ which is a $ 2 \times 2 $ matrix, which can be computed conveniently in a closed form.  Moreover, in order to incorporate a small neighborhood of gradient samples per pixel, we employ a diagonal weighting matrix $ \W_k $, constructed using a separable normalized Gaussian kernel.

Following \cite{feng}, the eigenvector $ \boldsymbol\phi^k_1 $, corresponding to the largest eigenvalue of $ \G_k^T\W_k\G_k $, can be used to derive the angle of the gradient $ \theta_k $, given by
\begin{equation}
\theta_k = \arctan(\boldsymbol\phi^k_{1,y}, \boldsymbol\phi^k_{1,x}). 
\end{equation}
Notice that due to the symmetry, a filter that corresponds to the angle $ \theta_k $ is identical to the one corresponding to $ \theta_k + 180^\circ $.


\begin{figure}
	\begin{center}
		\subfigure[$2 \times $ upscaling filters]{\includegraphics[width=0.85\textwidth]{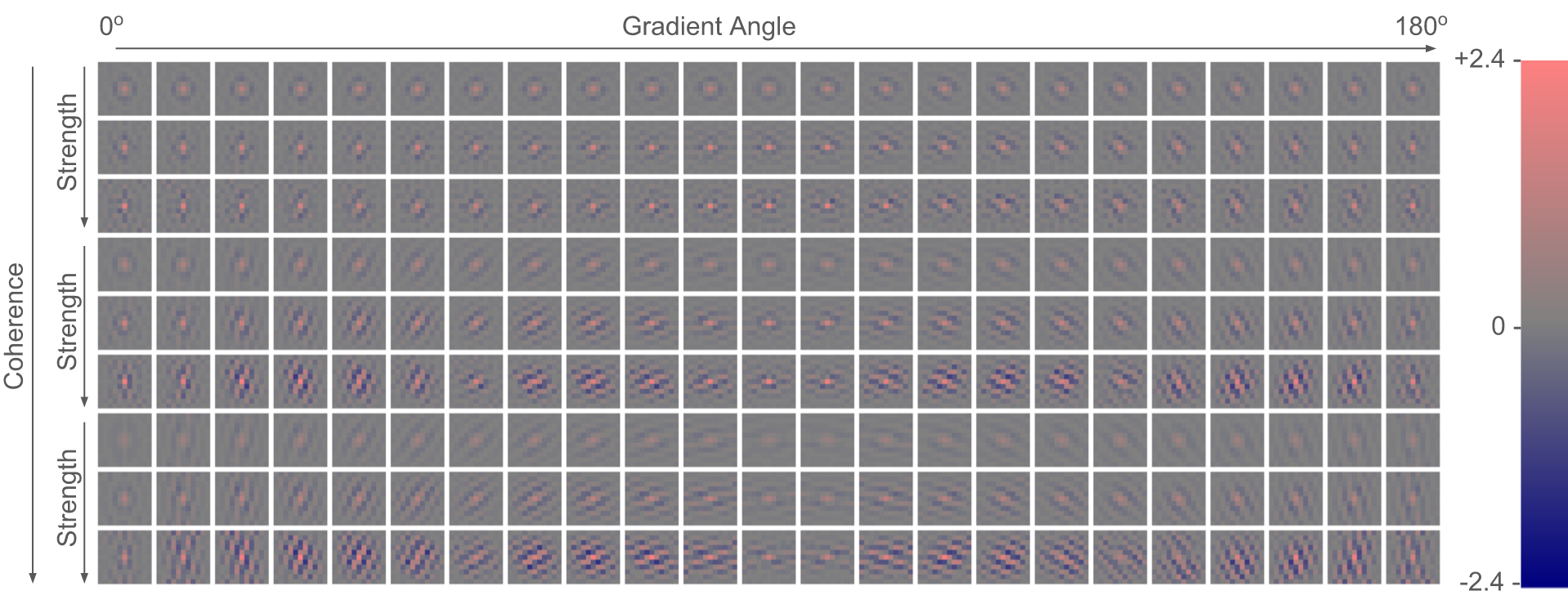}}		
		\subfigure[$3 \times $ upscaling filters]{\includegraphics[width=0.85\textwidth]{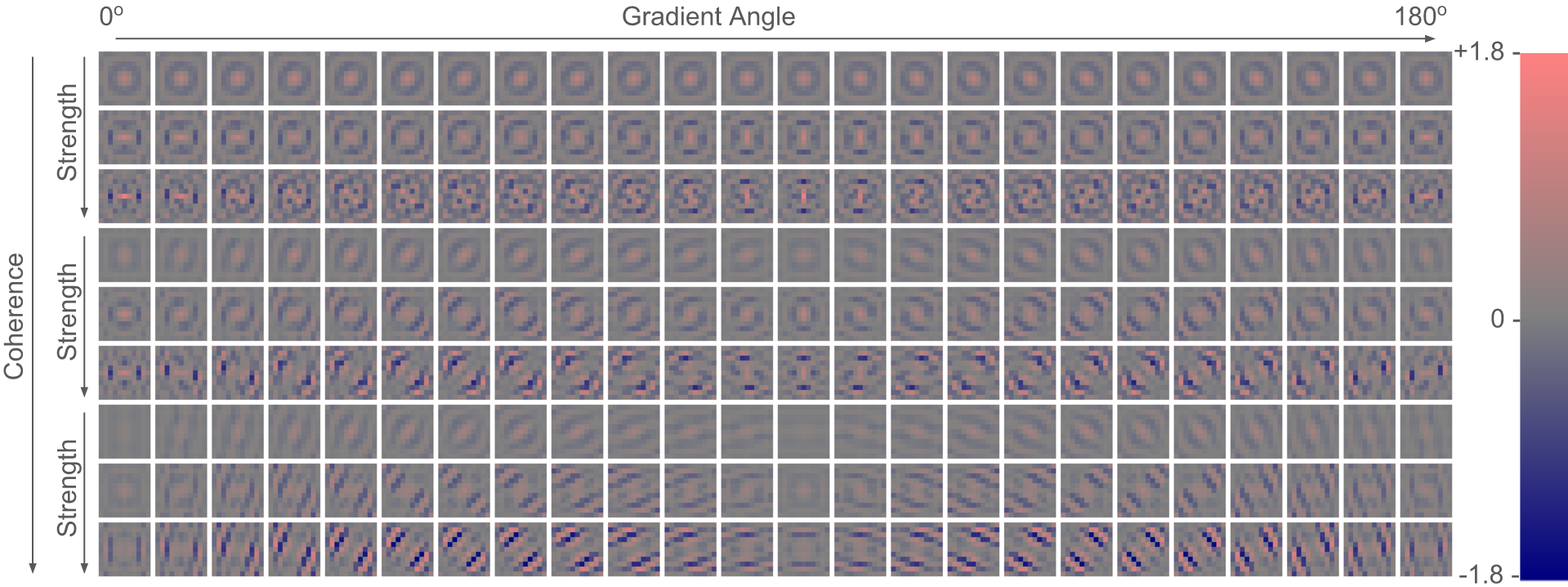}}
		\subfigure[$4 \times $ upscaling filters]{\includegraphics[width=0.85\textwidth]{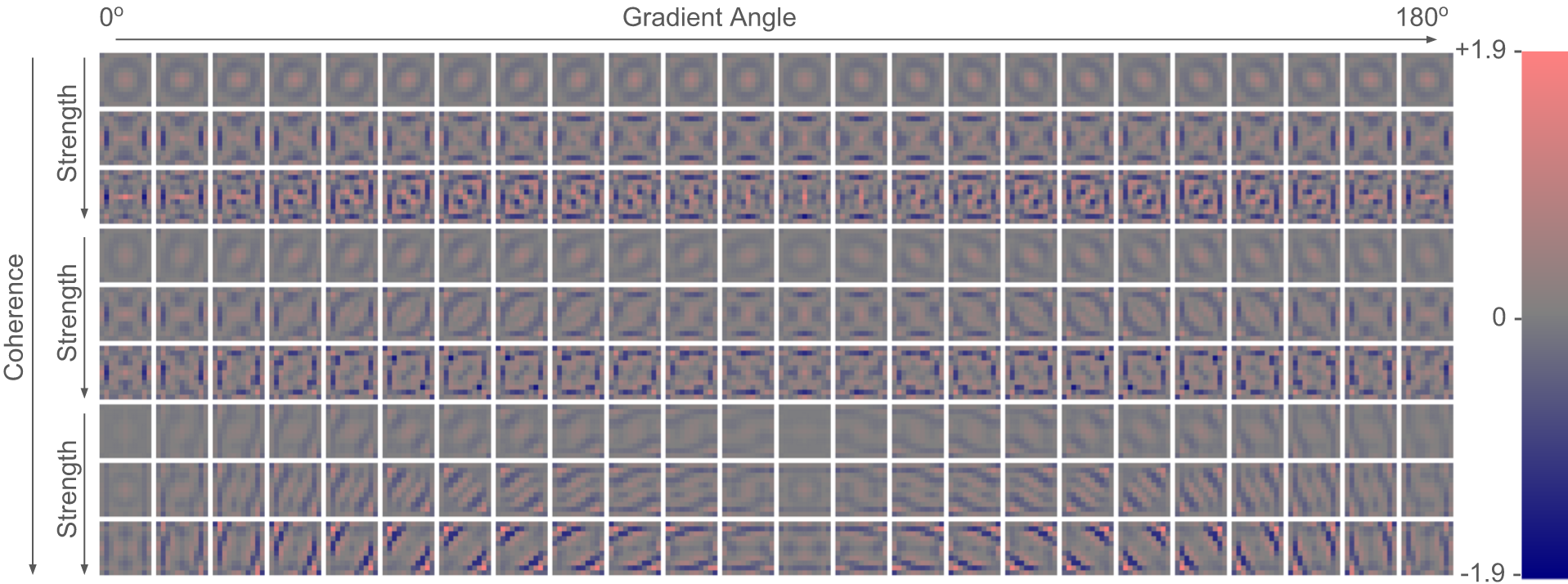}}
	\end{center}
	\caption{Visualization of the learned filter sets for (a) 2$ \times $, (b) 3$ \times $ and (c) 4$ \times $ upscaling, learned from using an angle, strength, and coherence based hashing scheme. Per each subset of filters, the angle varies from left to right; the top, middle, and bottom 3 rows correspond to low, medium and high coherence.  Within each set of 3 rows, gradient strength increases from top to bottom. As can be inferred, the general trend is that as coherence increases, the directionality of the filter increases. Also, as strength increases the intensity of the filter increases. Notice how the 3$ \times $ and 4$ \times $ upscaling filters are not simply scaled versions of the 2$ \times $ filters, but also have extracted additional information from the training data.}	
	\label{fig:oriented_filters_2x_3x_4x}
\end{figure}

As shown in \cite{feng}, the square root of the largest eigenvalue $ \lambda^k_1 $ is analogous to the ''strength'' of the gradient. The square root of the smaller eigenvalue $ \lambda^k_2 $ can be considered as the ''spread'' of the local gradients, or rather how much they vary in direction. Both of these can be measured in units of intensity.  The two eigenvalues can be combined into a unitless measure known as ''coherence'' \cite{feng}. The coherence value $ \mu_k $ ranges from 0 to 1, and formulated as
\begin{equation}
\mu_k=\frac{\sqrt{\lambda^k_1}-\sqrt{\lambda^k_2}}{\sqrt{\lambda^k_1}+\sqrt{\lambda^k_2}}.
\end{equation}

Strength and coherence are very useful for detecting a variety of different local image properties.  A low strength and low coherence often signifies a lack of image structure, and usually corresponds to noise or compression artifacts.  High strength, but low coherence often indicates corners or other multi-directional structure. Having a high coherence is generally an edge, or series of stripes in the same direction, with the strength measuring the relative intensity of the stripes. Intuitively, strength and coherence allow us to detect semantically different local image properties, so by using them as part of a hash enables the filter learning process to adapt to these conditions.  As such, combining angle $ \theta_k $, strength $ \lambda^k_1 $, and coherence $ \mu_k $  into a hash function, as detailed in Algorithm \ref{Alg:Hash}, can produce a family of learned filters that are able to handle a variety of situations.

\begin{algorithm}
	\caption{Computing the hash-table keys.}
	\textbf{Inputs}
	\begin{algorithmic}[1]
		\State Initial interpolated version of the LR image.
		\State $ Q_\theta $ -- Quantization factor for angle (e.g. 24).
		\State $ Q_s $ -- Quantization factor for strength (e.g. 3).
		\State $ Q_\mu $ -- Quantization factor for coherence (e.g. 3).
	\end{algorithmic}
	
	\textbf{Output}
	\begin{algorithmic}[1]
		\State Hash-table keys per pixel, denoted by $ \theta_k$, $ \lambda^k_1 $, and $ \mu_k $.
	\end{algorithmic}
	
	\vspace{0.2cm}
	
	\textbf{Process} \;
	\begin{itemize}[leftmargin=*]
	\item Compute the image gradients \;
	\item Construct the matrix $ \G_k^T\W_k\G_k $, and obtain the gradients' angle $ \theta_k $, strength $\lambda^k_1 $, and coherence $ \mu_k $ \;
	\item Quantize: 
	$ \theta_i \leftarrow \left \lceil \frac{\theta_i}{Q_\theta} \right \rceil$
	$ \lambda^i_1 \leftarrow \left \lceil \frac{\lambda^i_1}{Q_s} \right \rceil$ 
	$ \mu_i \leftarrow \left \lceil \frac{\mu_i}{Q_\mu} \right \rceil$, where $ \lceil \cdot \rceil $ is the ceiling function
	\end{itemize}
	\label{Alg:Hash}
\end{algorithm}

In Fig. \ref{fig:oriented_filters_2x_3x_4x}, one can see that bucketing by angle, coherence and strength produces a wide variety of filters. The ones that correspond to low coherency and strength tend to be bandpass and directionally invariant in nature.  
As coherency increases, so does the directionality of the filter, smoothing orthogonal to the gradient, but strongly sharpening in the direction of the gradient.


\subsection{Using Patch Symmetry for Nearly-Free 8$ \times $ More Learning Examples}
The amount of data needed for effective and stable learning of filter sets can be large. For instance, in practice, it takes at least $ 10^5 $ patches to reliably learn a given filter of size $ 9 \times 9 $ or $ 11 \times 11 $. Say we decompose the patches into $ \cal{B} $ buckets, when using a hashed set of filters, this implies that we need $ 10^5 $ patches per bucket. However, reaching this amount using real world training data is not as simple as using $ 10^5   \times  \cal{B} $ patches.

The problem emerges from the observation that certain hash values occur much more commonly than other hashes. There are often many more horizontal and vertical structures in imagery, and flat regions (such as sky, and painted surfaces) are common. Intuitively, these common structures result in the more common hashes.

\begin{figure}
	\begin{center}
		\includegraphics[width=\textwidth]{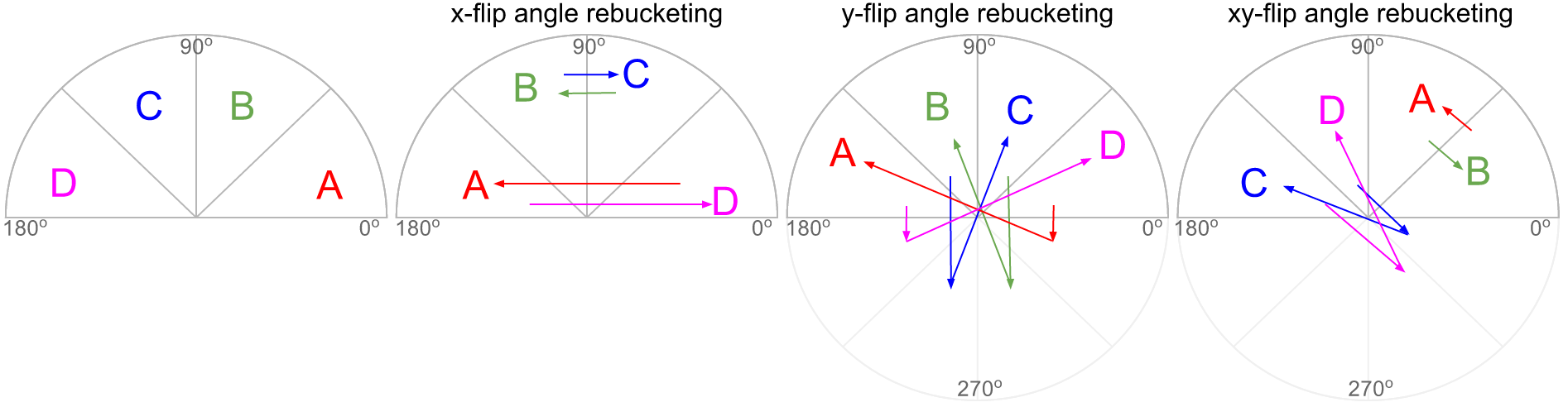}
	\end{center}
	\vspace{-0.2in}	
	\caption{Rebucketing for x-flips, y-flips, and xy-swaps. For example bucket 'B' accounts for gradient angles between $45^\circ$ and $90^\circ$. When the patch is flipped on the x-axis, the bucketing also needs to be reflected, as shown in the second graph. The third graph shows y-flips, for which the subsequent modulo $180^\circ$ operation maps back to the $0^\circ$ to $180^\circ$ range.  The final diagram shows xy-swaps (flip around the x$ = $y line).  The different possible combinations of x-flip, y-flip, and xy-swaps account for the 8 different patch transformations.  }
	\label{fig:SymmetryRebucketing}
\end{figure}

In order to make the patches that hash to uncommon hash values more effective, we leverage the patch symmetry and increase the learning power of each patch.
More specifically, we can generate $ 8 $ different example patches; four $ 90^\circ $ rotations, and four mirrored $ 90^\circ $ rotations.  Since $ 8 $ patches are generated for each original patch, we are effectively incorporating 8 times as much information for learning.  Since each transformation is a rotation and mirroring, the transformed patches often belong to a different hash bucket and shift.  For example, a patch that is rotated by $ 90^\circ $ changes the hash angle bucket by $ 90^\circ $. A visual demonstration of the proposed idea is given in \mbox{Fig. \ref{fig:SymmetryRebucketing}}, which demonstrates a simplified hashing scheme with 4 angular buckets on a polar graph, and how they get rebucketed from x-flips, y-flips, and xy-swaps.

Moreover, the patch transformations do not need to be performed as actual image transformations of each incoming patch, which would be expensive. If the gradient angle dependent hash bucket boundaries are symmetric to flips in x, y and xy-swaps, the accumulation for transformed patches can be performed.  Achieving this symmetry is a function of the hashing function, and, in our case, having a number of angle buckets that is divisible by 4 satisfies this requirement. As such, we can accumulate the per bucket and per pixel-type matrices $ \A^T_q\A_q $ and $ \A^T_q\b $ across all training samples, as suggested in Eq. \eqref{eq:compactLS_bucket}. Then, the symmetry can be applied as one last accumulation of permuted matrices, which act as a set of symmetry augmented matrices. In practice, the additional accumulation step to enable symmetry takes less than 0.1\% of the learning runtime (only a few additional seconds on a 3.4GHz 6-Core Xeon desktop computer).

\subsection{Built-in Suppression of Compression Artifacts and Sharpening Effect}
The linear degradation model that assumes blur and decimation, as expressed in Eq. \eqref{eq:degradation}, is very common in the literature, but less so in the real world. For example, often times the measured images are blurred with unknown kernel, compressed, post-processed (e.g. by applying gamma correction), contaminated by noise, and more. 

Learning a mapping that is capable to handle highly non-linear degradation model can be done by RAISR. We found that an effective suppression of compression artifacts is achieved by learning a mapping from \emph{compressed} LR images to their uncompressed HR versions. Notice that an inherent compression parameter is the bit-rate/compression quality, affecting the outcome of the learning scheme. For example, JPEG encoders use a quality level parameter, which varies from 0 (the worst quality) up to 100 (the best quality). Our experiments show that an aggressive compression (e.g. 80) indeed suppresses the compression artifacts, but can lead to smoothed result. We also found that a moderate compression level (e.g. 95) in training helps to suppress the aliasing in addition to alleviating moderate compression artifacts. 

In the same spirit, gaining sharpening effect can be done by learning a mapping from LR training images to their sharpened HR versions. The stronger the sharpening during the training phase, the sharper the outcome of RAISR upscaling. We should emphasize that at runtime, we simply apply the pre-learned filters (with possibly built-in sharpening effect); we do not apply a separate sharpening step. 

To conclude,   
by applying compression and sharpening as pre-processing steps, the learned filters are capable to map input compressed LR image to a sharpened HR output. As such, by choice, RAISR not only estimates the missing spatial information, but also suppresses the compression artifacts and amplifies the underlying signal.

\subsection{Blending: An Efficient Structure-Preserving Solution}
\label{sec:blending}
The proposed learning scheme results in adaptive upscaling filters, having built-in suppression of compression artifacts and sharpening effect. Concentrating on the sharpening property, two well known side-effects that exist are halos that appear along edges and the noise amplification. Put differently, applying the pre-learned filters on the initial interpolated image can lead to structure deformations due to the sharpening property.

As a way to avoid a significant modification in structure, we suggest measuring the local change in structure that occurred due to the filtering step, and blend accordingly. In areas that the structure of the initial interpolated image and the filtered one is somewhat similar -- we choose the filtered version, while in areas that a major change is occurred by the filtering step -- we choose the initial upscaled image. This suggestion relies on the observation that the cheap interpolated image typically does an adequate job on areas of the image that contain low spatial frequencies (e.g. flat regions). On the other hand, the higher spatial frequencies that need to be reconstructed require the careful treatment of the estimated filters. The blending therefore combines the most appropriate contributions from the cheap upscaled and the RAISR filtered image to yield the final result. One could have identified these regions ahead of time by clustering and apply different treatment, however this would have resulted in slower execution. In what follows, we present a fast alternative that works on two output images for point-wise blending.

\begin{figure}
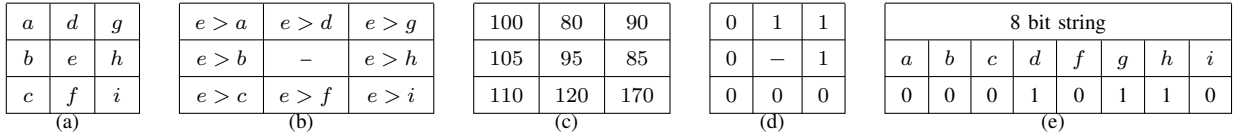

	\begin{center}
		\subfigure[]{
			\smaller
			\begin{tabular}{|c|c|c|}
				\hline
				$ a $ & $ d $ & $ g $ \\ \hline
				$ b $ & $ e $ & $ h $ \\ \hline
				$ c $ & $ f $ & $ i $ \\ \hline
			\end{tabular}
		}
		\hfill
		\subfigure[]{
			\smaller
			\begin{tabular}{|c|c|c|}
				\hline
				$ e > a $ & $ e>d $ & $ e>g $ \\ \hline
				$ e>b $ & -- & $ e>h $ \\ \hline
				$ e>c $ & $ e>f $ & $ e>i $ \\ \hline
			\end{tabular}
		}
		\hfill
		\subfigure[]{
			\smaller
			\begin{tabular}{|c|c|c|}
				\hline
				$ 100 $ & $ 80 $ & $ 90 $ \\ \hline
				$ 105 $ & $ 95 $ & $ 85 $ \\ \hline
				$ 110 $ & $ 120 $ & $ 170 $ \\ \hline
			\end{tabular}
		}
		\hfill
		\subfigure[]{
			\smaller
			\begin{tabular}{|c|c|c|}
				\hline
				$ 0 $ & $ 1 $ & $ 1 $ \\ \hline
				$ 0 $ & $ - $ & $ 1 $ \\ \hline
				$ 0 $ & $ 0 $ & $ 0 $ \\ \hline
			\end{tabular}
		}
		\hfill
		\subfigure[]{
			\smaller
			\begin{tabular}{|c|c|c|c|c|c|c|c|}
				\hline
				\multicolumn{8}{|c|}{8 bit string} \\ \hline
				$ a $  & $ b $  & $ c $  & $ d $  & $ f $  & $ g $  & $ h $ & $ i $ \\ \hline
				0  & 0  & 0  & 1  & 0  & 1  & 1 & 0 \\ \hline
			\end{tabular}
		}
	\end{center}
	\vspace{-0.1in}
	\caption{Census transform. (a) 3x3 window of pixels, (b) Boolean comparisons between the center pixel and its neighbors, (c)-(e) Numerical example: (c) Intensity values, (d) The outcome of the boolean comparisons, (e) Census result, an 8 bit string that measures the local structure.}
	\label{fig:CT_example}
\end{figure}
	
Inspired by the CT descriptor \cite{zabih1994non}, we suggest using its outcome as an engine that detect structure deformations and revert the errors of the upscaler. In order to understand the blending mechanism, a brief overview about the CT is given here. This transform maps the intensity values of the pixels within a small squared region (e.g. of size $ 3\times 3 $) to a bit string that captures the image structure. The CT is based on the relative ordering of local intensity values, and not on the intensity values themselves. Following \mbox{Fig. \ref{fig:CT_example}}, the center pixel's intensity value (central cell in Fig. \ref{fig:CT_example}a and \ref{fig:CT_example}c) is replaced by a bit string descriptor of length $ 8 $ (Fig. \ref{fig:CT_example}e), composed of a set of boolean comparisons between the center pixel and its $ 3\times 3 $ neighborhood pixels (Fig. \ref{fig:CT_example}b, \ref{fig:CT_example}d). Note that, in practice, when applying the comparisons we allow small variations in the intensity values, controlled by a threshold.  

\begin{figure}
	\begin{center}
		\includegraphics[width=5in]{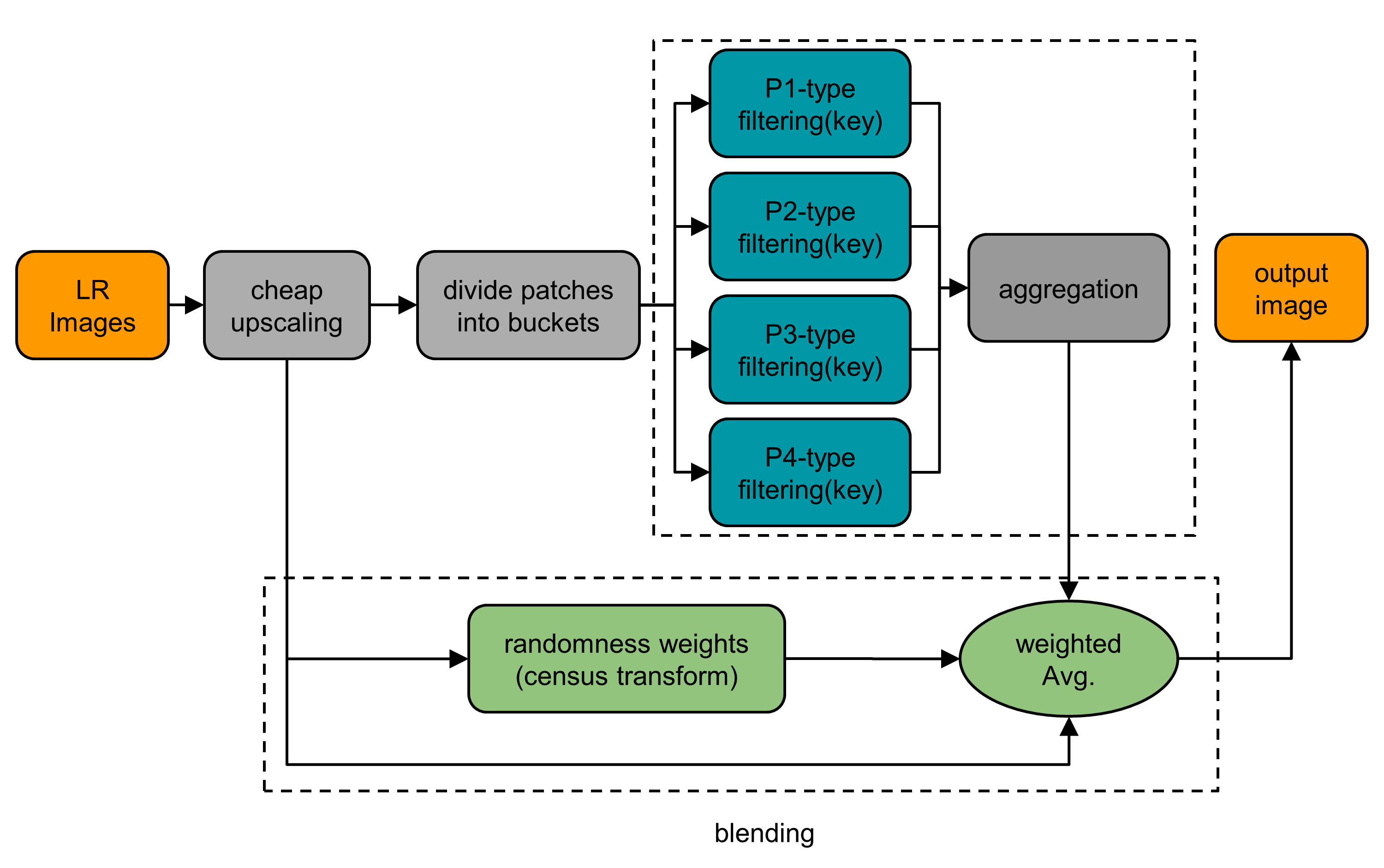}
	\end{center}
	\vspace{-0.3in}
	\caption{RAISR upscaling that allows the amplification of high-frequencies only. Applying the pre-leaned filters and avoiding halos and noise amplification. The blending select the filtered pixels in structured areas, and the cheap upscaled pixels in flat areas.}
	\label{fig:random}
\end{figure}

\begin{figure}
	\begin{center}
		\includegraphics[width=4.5in]{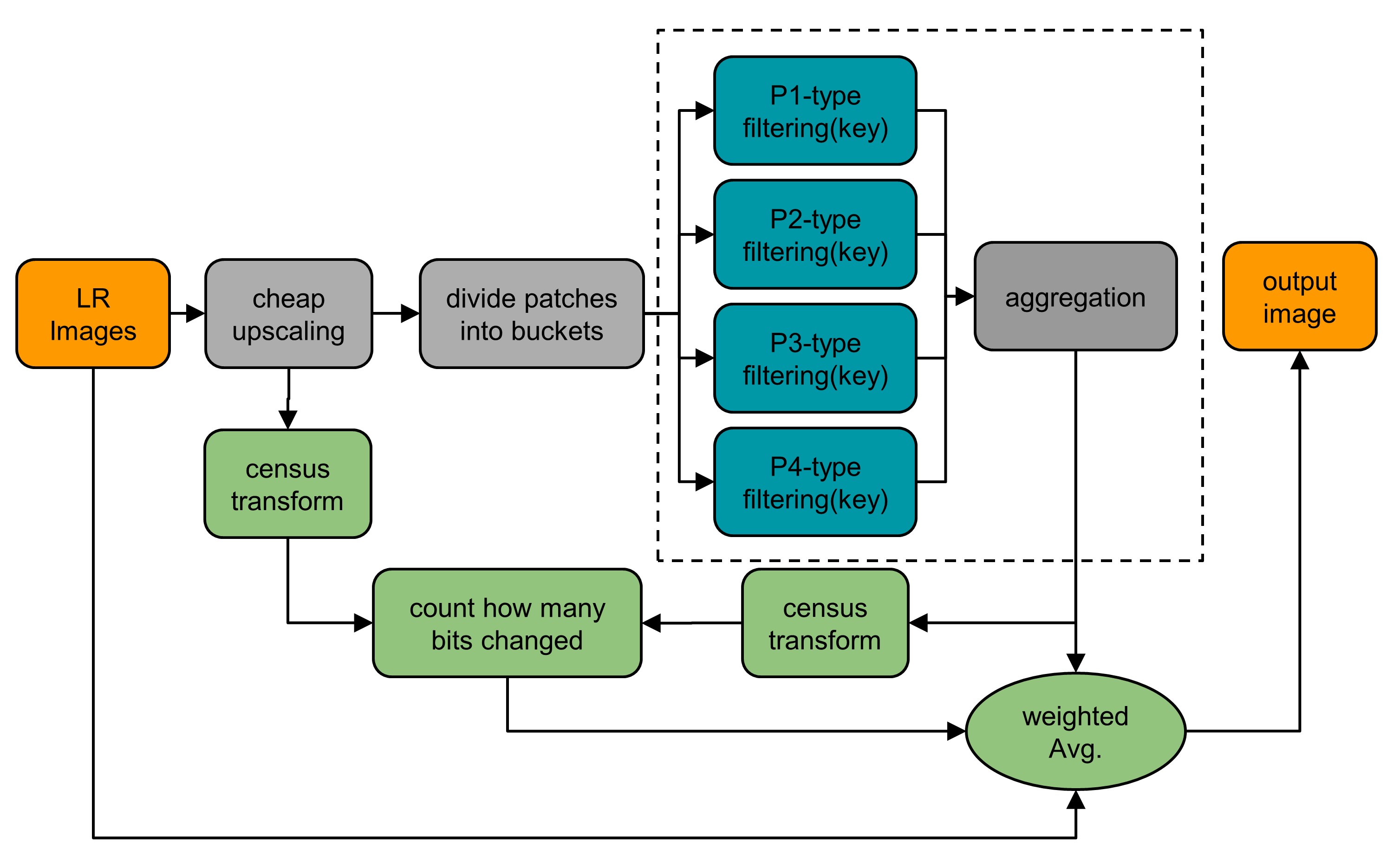}
	\end{center}
	\vspace{-0.15in}
	\caption{RAISR upscaling that allows the amplification of wide range of frequencies, enabling a contrast enhancement effect. Applying the pre-leaned filters and avoiding halos and noise amplification. The blending mechanism enables efficient structure and contrast enhancement, i.e., the amplification of a wide range of frequencies.}
	\label{fig:RAISR_CT}
\end{figure}

Back to RAISR, we suggest two different blending schemes, given in Fig. \ref{fig:random} and Fig. \ref{fig:RAISR_CT}, which result in different enhancement effects (as illustrated in Sections \ref{sec:sisr} and \ref{sec:all_in_one}, respectively). The first scheme allows the pre-learned filters to enhance only the high-frequencies of the image, leading to HR images that look natural, as required by the conventional SISR problem (a demonstration is given in Fig. \ref{fig:comp_vis}).
While the second scheme allows the enhancement of a wide range of frequencies, leading to better looking images due to the contrast enhancement effect (as shown in Fig. \ref{fig:comp_real}). Note that both versions aim at avoiding artifacts and structure deformations.

We start with the first blending mechanism, which enables the amplification of the high-frequencies only without modifying the low/mid frequencies, as usually done by the conventional SISR algorithms. Following Fig. \ref{fig:random}, our suggestion is based on the observation that in flat areas (or generally low frequency areas), a linear upscaler produces good results because there are no fine details to be recovered or aliasing to be suppressed, thus there is no need to further improve the results in these areas. On the other hand, the linear interpolation fails to recover structured areas, where the proposed pre-learned filters play the key role. Moreover, within the structured areas, especially along strong edges, the pre-learned filters may introduce halos due to their size ($ 11 \times 11 $ or $ 9 \times 9 $) and the sharpening property. 

For the sake of completeness, let us explain briefly how the CT, which is indeed blind to the illumination, can be formulated as a mechanism for edge/structure-detection, thus allowing us to amplify only the high-frequencies of the image. In this case, the blending weights are the outcome of the so called "randomness" measure, which indicates how likely a pixel is in a structured area. 
Specifically, the size of the Least Connected Component (LCC) of the CT descriptor (in the case of Fig. \ref{fig:CT_example}d the LCC size is 3) is translated to a weight, determining the strength/amount of the structure within the descriptor window. In general, the larger the size of LCC the higher the weight. Put differently, by measuring the ''randomness'' of the bit string we can infer whether the pixel is a part of an edge or not, forming the blending weights map. A block diagram of the proposed upscaling scheme, which allows sharpening of high-frequencies only, is given in Fig. \ref{fig:random}.

Although the conventional SISR algorithms result in HR images that look natural, contrast enhancement (i.e. amplification of low, mid, and high frequencies) often times can lead to better looking images (a visual demonstration is given in Fig. \ref{fig:comp_real}). This observation leads us to propose the second CT-based blending scheme. According to Fig. \ref{fig:RAISR_CT}, in order to measure the local change in structure, we
\begin{enumerate}[label = (\roman*)]
	\item Compute the CT of the initial upscaled image and the filtered image, and then
	\item Per each pixel -- count the number of bits that were changed, i.e., evaluate the Hamming distance, where the larger the distance, the larger the change in structure.
\end{enumerate}
As a consequence, by translating the number of bits that were changed to weights, we form the desired blending map. Notice that the CT is blind to the intensity value itself. Therefore, differently from the randomness strategy (see Fig. \ref{fig:RAISR_CT}), in this case, the obtained blending map allows a local change in the intensity (or contrast), while avoiding major changes in the structure. 

In the learning stage, the target HR images are pre-processed by the proposed DoG sharpener (a detailed explanation about the sharpener is given in Section \ref{sec:sharpener}), which sharpens the structures and improves the overall contrast of the image. As a result, a built-in enhancement of both details (high-frequencies) and contrast (mid/mid-low frequencies) is achieved when applying the pre-learned filters. A block diagram of the upscaling scheme that allows contrast enhancement (i.e. amplification of a wide range of frequencies) is shown in \mbox{Fig. \ref{fig:RAISR_CT}}.

Our experiments show that when enhancing only the high-frequencies we obtain images that look natural, having the same contrast of the LR ones (see Fig. \ref{fig:comp_vis}). When enhancing a wider range of frequencies (i.e. allowing contrast change), RAISR produces better looking images (as illustrated in Fig. \ref{fig:comp_real}), but there is no guarantee that the output will have the same nature of the one of the LR. If we use the versions that clean compression artifacts, do sharpening and contrast enhancement, the effect on the PSNR or SSIM comparisons will not be very clear anymore. So in terms of this quantitative measure, we might observe deterioration, even if the images look excellent (even better than the originals!).

To conclude, we introduced the RAISR algorithm that turns a LR image into a HR image. The process is carried out in several steps:
\begin{enumerate}[label=(\roman*)]
	\item A very cheap (e.g. bilinear) interpolation method is used to upscale the LR image.
	\item A hash-table, containing set of filters, is learned from a training database, where the hash-table keys are a function of gradient properties. The filters are applied on the output of step (i) to improve its quality.
	\item The outputs of steps (i) and (ii) are selectively blended (with different weights at each pixel) to produce the final result.
\end{enumerate}
As a closing remark, the whole learning phase is summarized in Algorithm \ref{Alg:Learning} and the upscaling stage is detailed in Algorithm \ref{Alg:Upscaling}.

\begin{algorithm}		
	\caption{RAISR: Hashing-based learning.}
	
	\textbf{Inputs}
	\begin{algorithmic}[1]
		\State $ s $ -- Upscaling factor.		
		\State $ \{\x_i\}_{i=1}^{L} $ -- The ground truth HR images (optionally sharpened/enhanced).
		\State $ \{\z_i\}_{i=1}^{L} $ -- The LR (optionally compressed) versions of $ \{\x_i\}_{i=1}^{L} $. 
	\end{algorithmic}
	
	\textbf{Output}
	\begin{algorithmic}[1]
		\State $ \mathcal{H}_t : \left( j \right) \mapsto \h_{j,t}$ -- Hash-table that maps each key $j = (\theta, \lambda_1, \mu )$ to the corresponding $ d^2 $-dimensional filter $ \h_{j,t} $, where $ 1 \leq t \leq s^2 $ is the pixel/patch type.		
	\end{algorithmic}
	
	\vspace{0.2cm}
	
	\textbf{Process} \;
	\begin{itemize}[leftmargin=*]
	\item $  \Q_{j,t} \leftarrow \mathbf{0}, \ \V_{j,t} \leftarrow \mathbf{0} $ for all hash-table keys denoted by $ j $ and pixel-type $ t $\;
	\end{itemize}
	\For{$i=1$ \emph{\KwTo} $L$} {
		\begin{itemize}[leftmargin=*]
		\item Compute $ \y_i $, an initial interpolated version of $ \z_i $ \;
		\item Initialize $  \A_{j,t}$ and $\b_{j,t}$ to be empty matrices for all possible hash-table keys denoted by $ j $ and pixel-type $ t $\;
		\end{itemize}
		\For{\text{each pixel $ k $ in $ \y_i $}} {
			\begin{itemize}[leftmargin=*]
			\item Denote by $ j = (\theta_k,\lambda_1^k,\mu_k) $ the hash-table key of the pixel $ k $ \;
			\item Denote by $ \p_k \in \RR^{d^2}$ the patch extracted from $ \y_i $ centered at $ k $ \;
			\item Denote by $ \x_i(k) $ the ground truth HR pixel of $ \x_i $ located at $ k $ \;
			\item Denote by $ t $ the type of the pixel $ k $ \;
			\item $ \A_{j,t} \leftarrow [\A_{j,t} \ ; \ \p_k^T]$, i.e., concatinate $ \p_k^T $ to the end of $ \A_{j,t} $ \;
			\item $ \b_{j,t} \leftarrow [\b_{j,t} \ ; \ \x_i(k)]$, i.e., concatinate $ \x_i(k) $ to the end of $ \b_{j,t} $ \;
			\end{itemize}
		}
		\For{\text{each key $ j $ and pixel-type $ t $}} {
			\begin{itemize}[leftmargin=*]			
			\item $ \Q_{j,t} \leftarrow \Q_{j,t} + \A_{j,t}^T\A_{j,t}$ \;
			\item $ \V_{j,t} \leftarrow \V_{j,t} + \A_{j,t}^T\b_{j,t}$ \; 
			\end{itemize}
		}
		
	}
	\For{\text{each key $ j $ and pixel-type $ t $}} {
		\begin{itemize}[leftmargin=*]
		\item $\h_{j,t} \leftarrow \arg\min_{\h}  \| \Q_{j,t} \h -\V_{j,t} \|_2^2 $ \;
		\item $ \mathcal{H}_t\left( j \right) \leftarrow \h_{j,t} $ \;
		\end{itemize}
	}
	
	\label{Alg:Learning}
\end{algorithm}

\begin{algorithm}
		
	\caption{RAISR: Hashing-based upscaling.}
	\textbf{Inputs}
	\begin{algorithmic}[1]
		\State $ s $ -- Upscaling factor.
		\State $ \z $ -- LR image.
		\State $ \mathcal{H}_t : \left( j \right) \mapsto \h_{j,t}$ -- Hash-table that maps each key $j = (\theta, \lambda_1, \mu )$ to the corresponding $ d^2 $-dimensional filter $ \h_{j,t} $, where $ 1 \leq t \leq s^2 $ is the pixel/patch type.	
	\end{algorithmic}	
	
	\textbf{Output}
	\begin{algorithmic}[1]
		\State $ \xh $ -- HR estimate of $ \z $.
	\end{algorithmic}
	
	\vspace{0.2cm}
	
	\textbf{Process} \;
	\begin{itemize}[leftmargin=*]
	\item Compute $ \y $, an initial interpolated version of $ \z $ \;
	\end{itemize}
	\For{ \text{each pixel $ k $ in $ \y $}} {
		\begin{itemize}[leftmargin=*]		
		\item Denote by $ j = (\theta_k,\lambda_1^k,\mu_k) $ the hash-table key of the pixel $ k $ \;
		\item Denote by $ \p_k \in \RR^{d^2}$ the patch extracted from $ \y $ centered at $ k $ \;
		\item Denote by $ t $ the type of the pixel $ k $ \;
		\item $ \h_{j,t} \leftarrow \mathcal{H}_t\left( j \right) $ \;
		\item $ \xh(k) \leftarrow \p_k^T\h_t$ \;
		\end{itemize}
	}
	\begin{itemize}[leftmargin=*]
	\item $ \xh \leftarrow \text{CT-Blending}(\xh, \y) $
	\end{itemize}
	 
	\label{Alg:Upscaling}
\end{algorithm}

\section{CT-Based DoG Sharpener}
\label{sec:sharpener}
In this section we briefly revisit the Difference of Gaussian (DoG) operator, which is widely used for edge enhancement \cite{winnemoller2012xdog}. Then, we introduce a very efficient way to sharpen an image using this operator while eliminating halos, noise amplification, and other similar common sharpening artifacts.

The DoG filter is formulated as a subtraction of two low-pass filters -- Gaussians with different standard-deviation $ \sigma $. In general, a Gaussian filter attenuates the high frequencies of the image, where the parameter $ \sigma $ controls the blurring effect, i.e., modifies the cut-off point of the filter. As such, when subtracting two Gaussians with different standard-deviations, we manually design a bandpass filter that reduces the amplitude of all frequencies between two cut-off points. 
More formally, a DoG filter $ \mathcal{D}_{\sigma,\alpha,\rho}(\z) $ can be expressed by
\begin{equation}
\mathcal{D}_{\sigma,\alpha,\rho}(\z) = (1+\rho) \mathcal{G}_\sigma(\z) - \rho\mathcal{G}_{\alpha \sigma}(\z),
\end{equation}
where $ \mathcal G_\sigma(\z) $ is a Gaussian filter with standard deviation $ 
\sigma $, the scalar $ \rho $ controls the amplification factor, and $ \alpha $ is a constant that controls the range of the frequencies that we wish to pass. 
According to the chosen parameters, the filter $ \mathcal{D}_{\sigma,\alpha,\rho}(\z) $ captures different frequency-bands, thus being an efficient mechanism to amplify a wide range of frequencies by choice \cite{winnemoller2012xdog}.


\begin{figure}
	\begin{center}
		\includegraphics[width=4in]{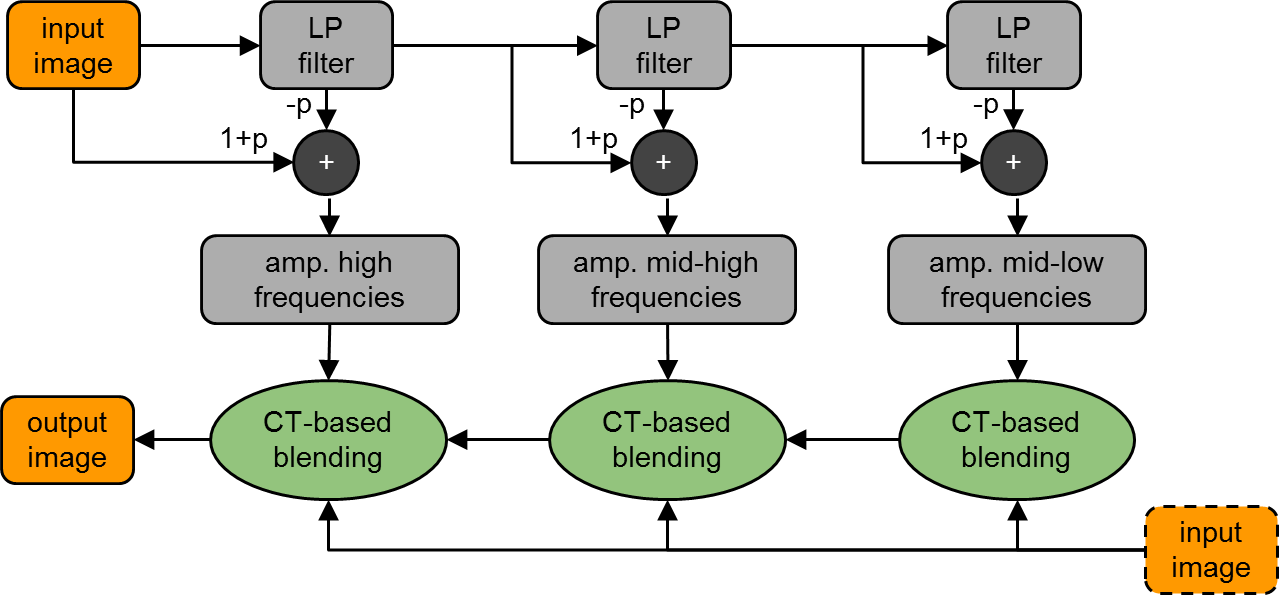}
	\end{center}
	\vspace{-0.2in}
	\caption{CT-based DoG sharpener: Applying the DoG filters on the image, followed by blending steps that utilize the outcome of the CT.}
	\label{fig:CT_DoG}
\end{figure}

Similarly to the explanation that is given in Section \ref{sec:blending}, when dealing with sharpening two main issues are raised: (i) Noise amplification, and (ii) Halos artifacts. The noise may exist in the whole frequency domain, thus amplified by the naive DoG sharpener along with the underlying signal. In addition, since this sharpener is not adaptive to the content/structure of the image it tends to produce sharpening artifacts, such as halos, over-sharpening, gradient-reversals and more. Differently from the linear DoG filter, the content-aware non-linear filtering methods \cite{zhang2008adaptive,he2010guided,talebi2014nonlocal,kheradmand2015nonlinear,liu2015joint} successfully avoid these common artifacts. However, computationally, they are more complex than the linear approach. 

We wish to keep the computational advantages of working with the linear DoG filter, while gaining adaptivity to the content of the image. Similarly to RAISR, we use the ''blending trick'' (see Section \ref{sec:blending}) once again. By integrating the CT blender to the naive DoG sharpener scheme we achieve the desired content-aware property, in an extremely efficient fashion. The blending is similar to the one that described in the context of RAISR, with the exception that now there are several images that are fused together. Specifically, the input image is locally blended with its different enhanced versions, obtained by applying naive DoG filters that amplifies different frequency-bands. 

Moreover, in order to reduce computations, we suggest a cascade implementation that utilizes the already filtered images between the levels (see Fig. \ref{fig:CT_DoG}). For example, instead of applying a wide separable Gaussian filter in order to capture the mid-low frequency band, one can filter the already computed smoothed image, obtained in the previous cascade level. As can be easily inferred, the complexity of the proposed sharpener is very low; it is equivalent to the application of linear separable filters on the image, followed by a pixel-wise weighted average. Notice that the weights are a function of very cheap descriptor that applies extremely basic manipulations on $ 3\times3 $ neighborhood of a pixel (e.g. boolean comparisons and evaluation of Hamming distance).

Following Fig. \ref{fig:CT_DoG}, notice that different blending mechanisms leads to different enhancement effect: When counting the modified bits of the CT descriptor, we are capable to enhance the contrast of the image (even in relatively low-frequencies). On the other hand, when choosing the randomness measure as a blending map, we enhance the content  only along edges and structures.

\section{Experiments}
\label{sec:experiments}

In this section we test the proposed algorithm in various scenarios. First, in the context of the conventional SISR, the effectiveness of RAISR is tested and compared to several state-of-the-art algorithms for $ 2\times $, $ 3\times $, and $ 4\times $ upscaling. Then, the abilities of the proposed algorithm to tackle real-world scenarios are demonstrated by applying RAISR upscaling on arbitrary compressed images. In this case, we show that RAISR is able to produce contrast-enhanced high quality images by learning filters from compressed LR images to their contrast-enhanced HR versions. The pre-processing of the training images is done by applying the proposed CT-based DoG sharpener.

\subsection{Single Image Super-Resolution}
\label{sec:sisr}

In this subsection we compare the performance of proposed algorithm with several state-of-the-art methods for $ 2\times $, $ 3\times $, and $ 4\times $ upscaling factors, both quantitatively and visually. All the results are obtained by applying the scheme that is given in Fig. \ref{fig:random} (and its variants for $ 3\times $ and $ 4\times $ upscaling), followed by a back-projection step (see Appendix \ref{app:back}). The filters are learned using a collection of 10,000 advertising banner images. This imagery type was chosen for its wide variety of both synthetic and real world content.

For all upscaling factors, we learn filters of size $ 11\times11 $. In the context of the hash-table, we consider a neighborhood of size $ 9\times9 $ for the computation of the angle, strength, and coherence of the gradients. Also, the quantization factors of the angle $ Q_\theta $, strength $ Q_s $, and coherence $ Q_\mu $ are set to result in a total of 216 total buckets (24 angular bins by 3 strength bins by 3 coherence bins). In addition, we found that an amplification of the high-frequencies of the training-set HR images leads to an improved restoration. The proposed CT-based DoG sharpener (see Fig. \ref{fig:CT_DoG}) is utilized for this task, where we use the randomness measure as the blending map; the number of layers is set to 3, with $ \sigma = 0.85 $, $ \alpha = \sqrt{2} $ and $ \rho = 55$. Note that a fast approximation of separable Gaussian filter is suggested, obtained by applying one dimensional $ [1, \ 2, \  1]/4 $ filter in a separable fashion (acts as a LP Gaussian filter with $ \sigma\approx0.85 $).

As done by the leading algorithms \cite{timofte2014a+,dong2014learning,dong2015}, the proposed algorithm is tested on 2 widely used databases: (i) Set5 \cite{bevilacqua2012low}, and (ii) Set14 \cite{zeyde2012single}, which are composed of 5 and 14 standard images, respectively. For all scaling factors, the test LR images are generated by downscaling the original HR images using the bicubic interpolation.

The restoration performance is evaluated using the Peak Signal to Noise Ratio (PSNR), and Structural Similarity (SSIM) metrics, measured between the luminance channel of the original and the estimated images. The higher these measures the better the restoration. Note that upscaling an RGB image is done by converting it to the YCbCr color space. Then, a bicubic interpolation is applied on the luminance channel, which is used as an initialization for RAISR, while the chromatic channels are upscaled only by the bicubic interpolation. Finally, the estimated HR channels are converted back to the RGB color space. 

\begin{figure}
	\begin{center}
		\subfigure[PSNR, $2 \times $ upscaling]{\includegraphics[width=0.49\linewidth]{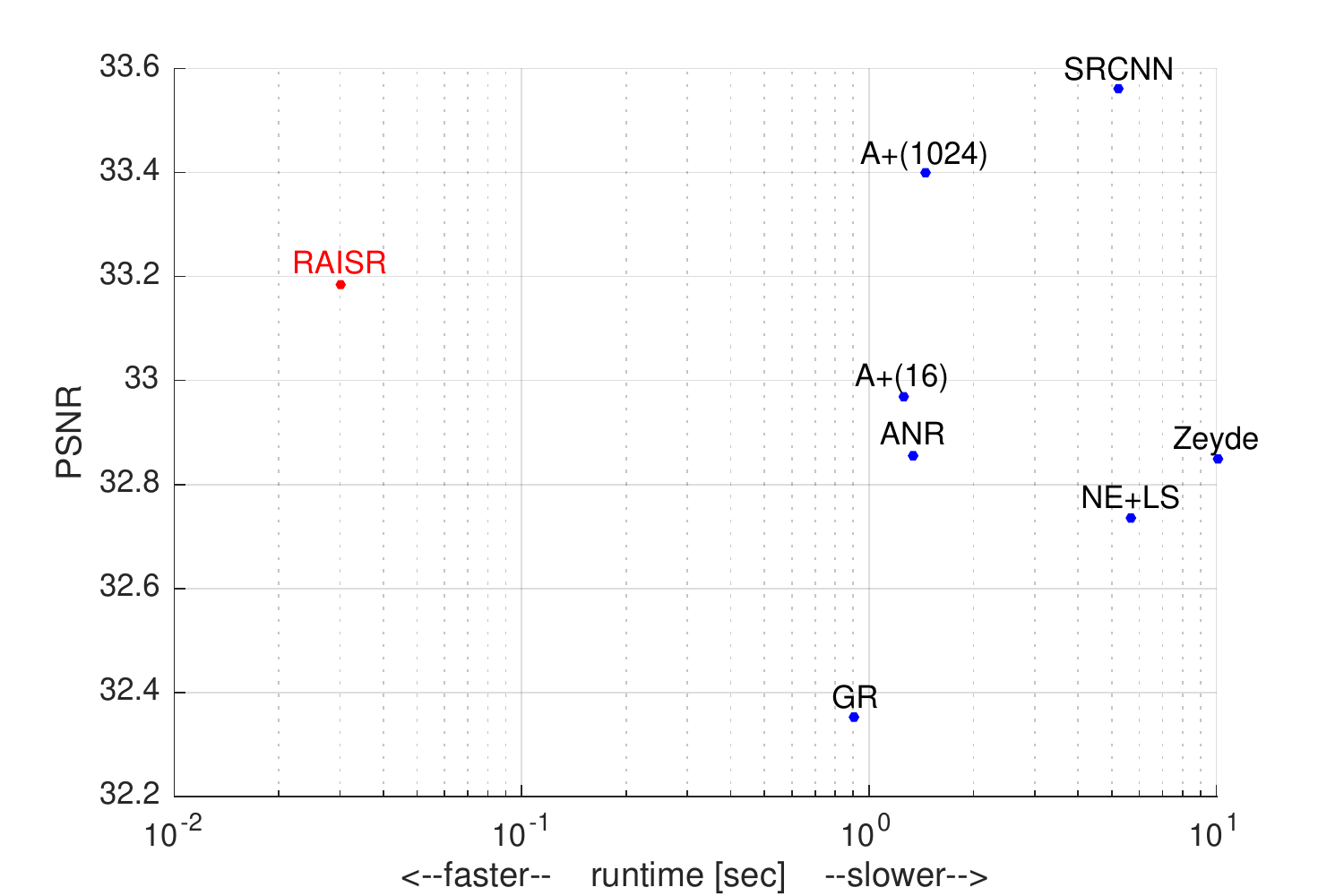}}
		\subfigure[SSIM, $2 \times $ upscaling]{\includegraphics[width=0.49\linewidth]{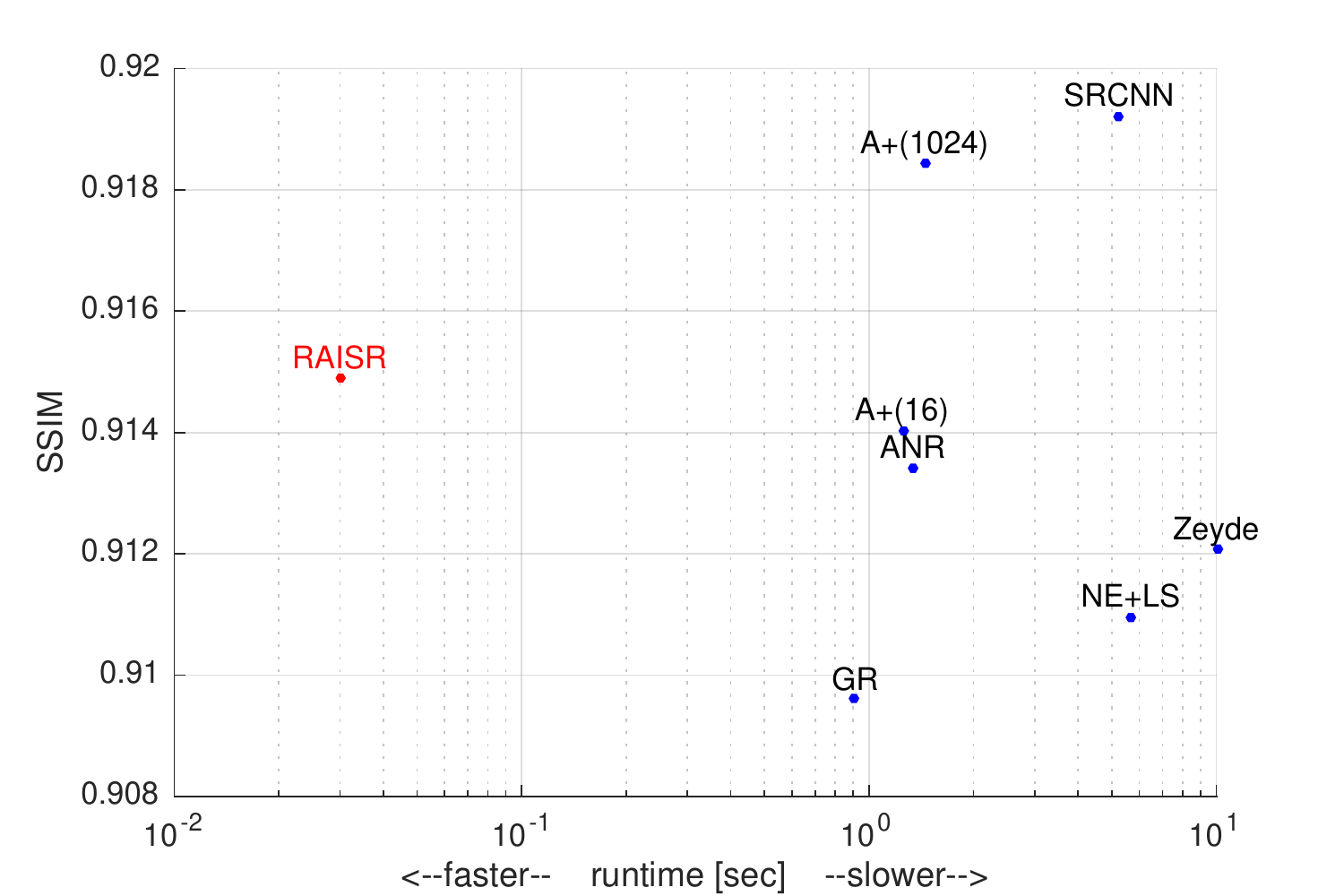}}
		\\
		\subfigure[PSNR, $3 \times $ upscaling]{\includegraphics[width=0.49\linewidth]{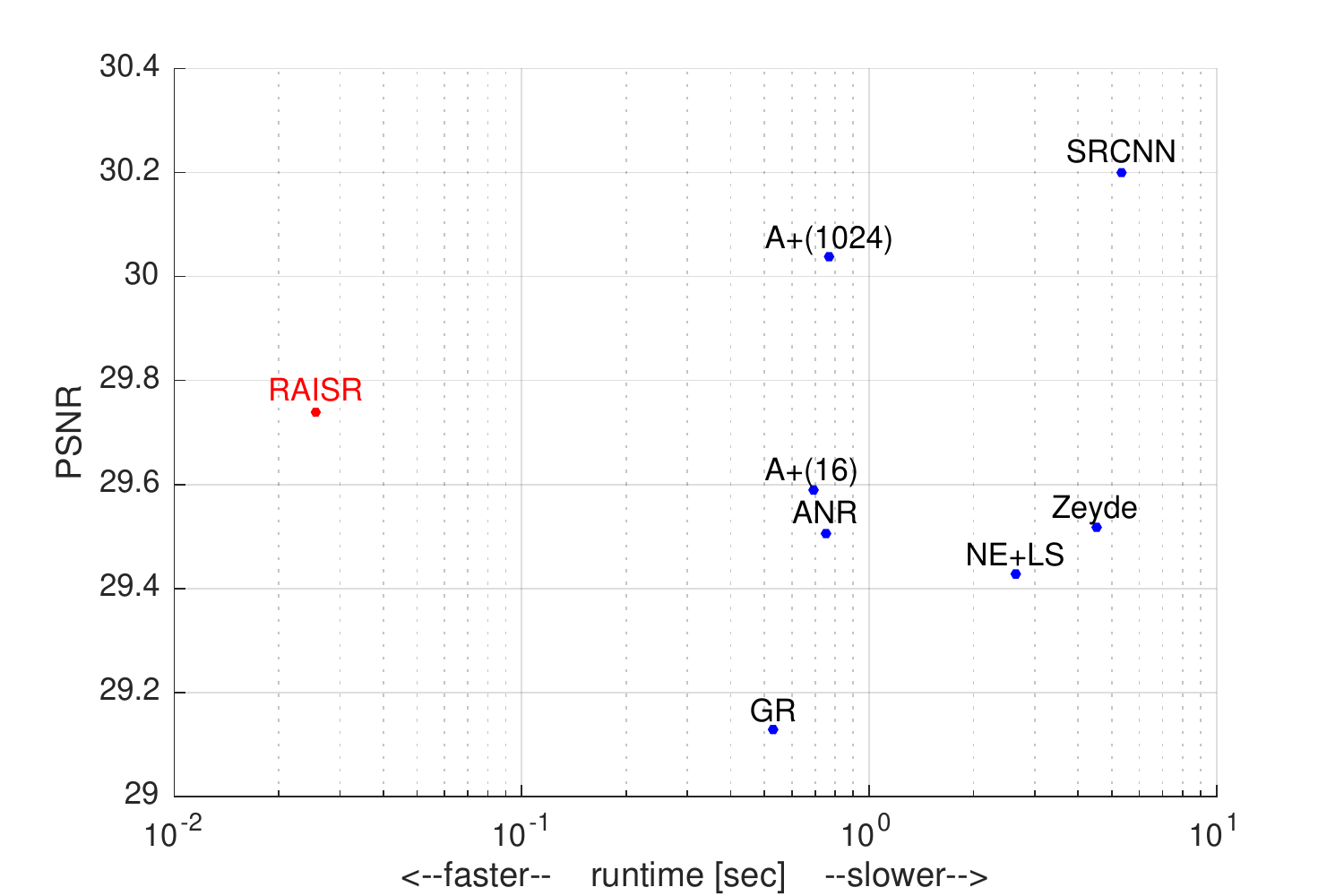}}
		\subfigure[SSIM, $3 \times $ upscaling]{\includegraphics[width=0.49\linewidth]{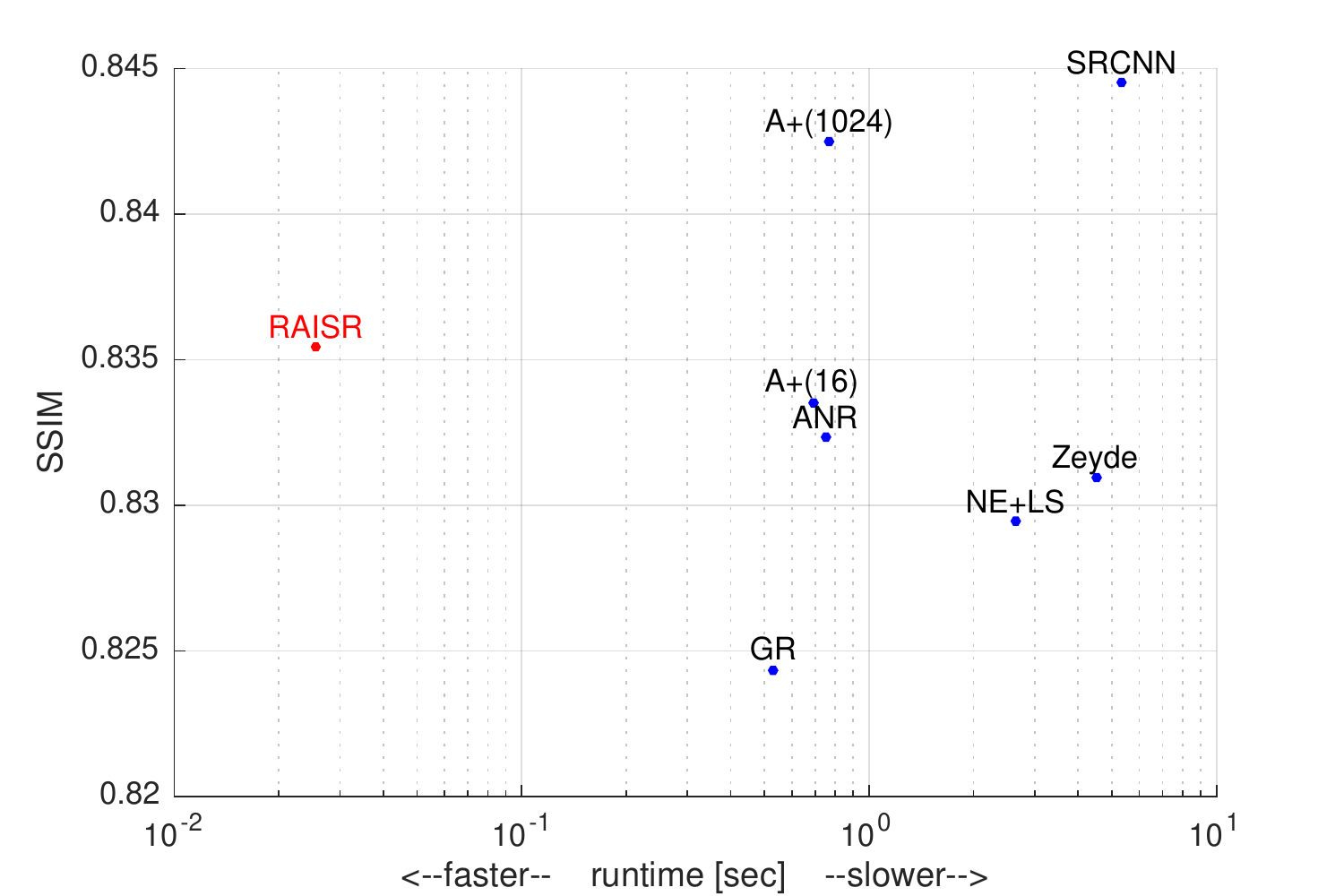}}
		\\
		\subfigure[PSNR, $4 \times $ upscaling]{\includegraphics[width=0.49\linewidth]{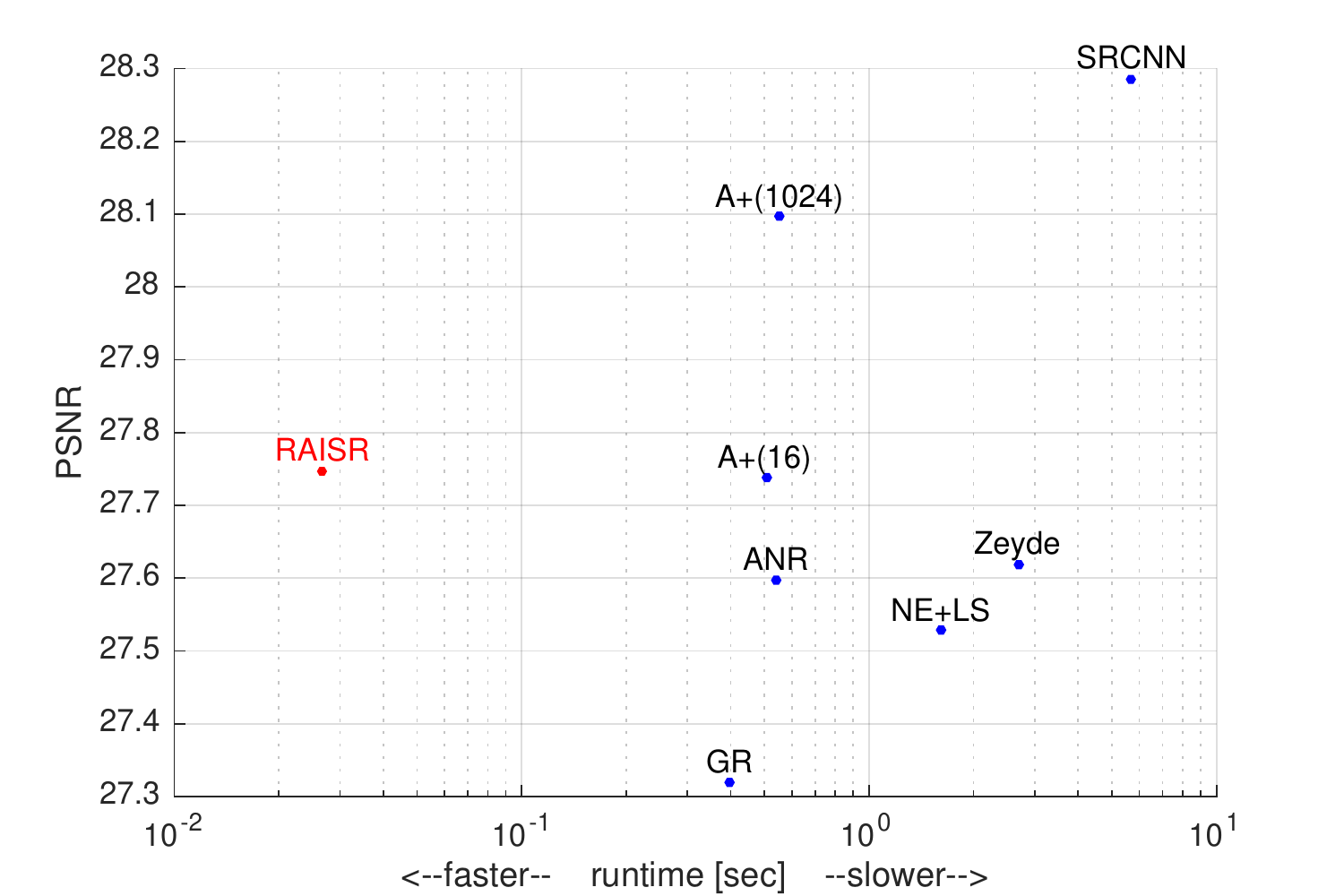}}
		\subfigure[SSIM, $4 \times $  upscaling]{\includegraphics[width=0.49\linewidth]{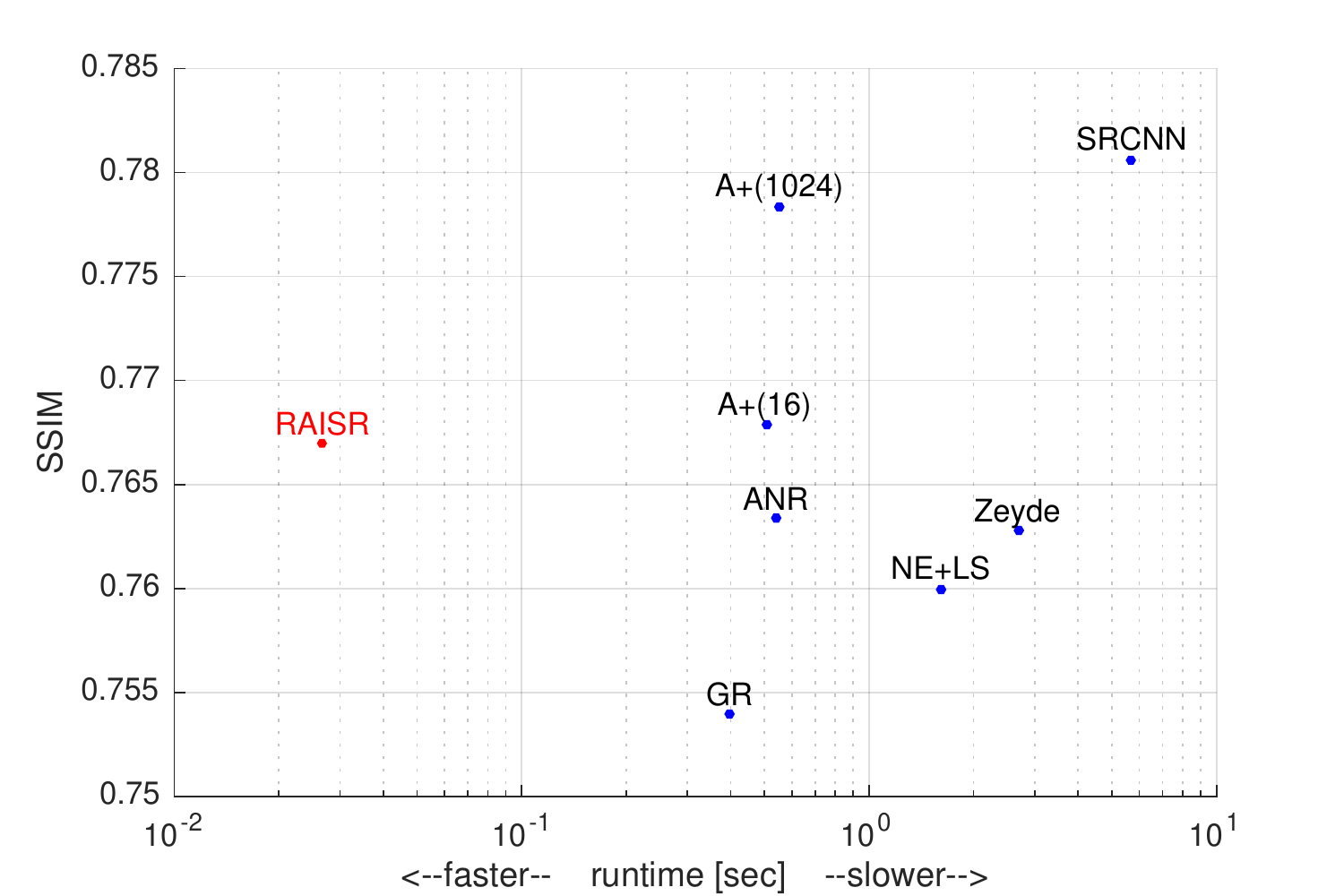}}
	\end{center}
	\vspace{-0.1in}
	\caption{Quantitative comparison between the restoration performance  vs. runtime. The vertical coordinate of each point in the scatter plots corresponds to the average PSNR (a,c,e) and SSIM (b,d,f) of each method, measured on the test images of Set5 and Set14. The horizontal coordinate corresponds to the average runtime (the average size of the upscaled images is about $ 0.63\times10^6 $ pixels). The size of each point reflects the standard error of the PSNR/SSIM. Detailed quantitative results can be found in the \href{https://goo.gl/D0ETxG}{supplementary material} (website address can be found at the bottom of the first page of this paper).}
	\label{fig:comp}
\end{figure}

The proposed algorithm is compared to various methods, including the sparse-coding approach of Zeyde et al. \cite{zeyde2012single}, along with the efficients GR and ANR \cite{ANR}. Since we put emphasis on runtime, we test two versions of A+ \cite{timofte2014a+}, the first is a fast version that uses 16 atoms, while the second produces state-of-the-art results (with the cost of increased runtime) by utilizing 1024 atoms. In addition, we compare our algorithm to the state-of-the-art SRCNN \cite{dong2014learning,dong2015}, which is based on a powerful convolutional neural network architecture. A neighbor embedding technique is also included, called NE+LLE, which assumes that the LR-HR patches lie on low-dimensional manifolds, having locally similar geometry \cite{chang2004super}\footnote{We would like to thank the authors of \cite{zeyde2012single,timofte2014a+,dong2014learning,dong2015} for providing the software that produces the results of NE+LLE, Zeyde et al., GR, ANR, A+, and SRCNN. In all cases we use the original codes, with default parameters.}. We should note that all the baseline methods significantly outperform the bicubic interpolation.

\begin{figure}
	\begin{center}
		\subfigure[Bicubic]{\includegraphics[width=0.24\linewidth]{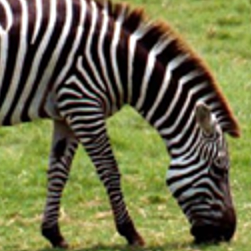}}
		\subfigure[SRCNN]{\includegraphics[width=0.24\linewidth]{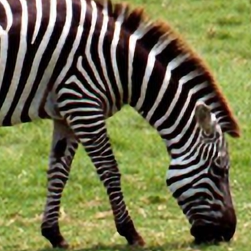}}
		\subfigure[A+]{\includegraphics[width=0.24\linewidth]{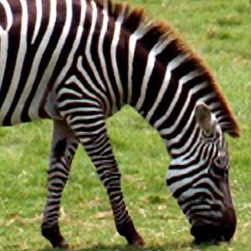}}
		\subfigure[RAISR]{\includegraphics[width=0.24\linewidth]{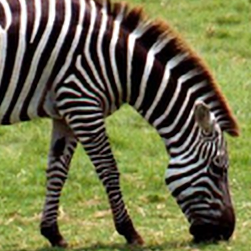}}
		\\
		\subfigure[Bicubic]{\includegraphics[width=0.24\linewidth]{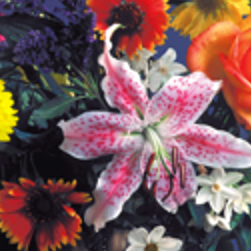}}
		\subfigure[SRCNN]{\includegraphics[width=0.24\linewidth]{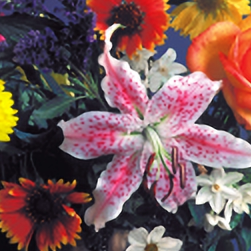}}				
		\subfigure[A+]{\includegraphics[width=0.24\linewidth]{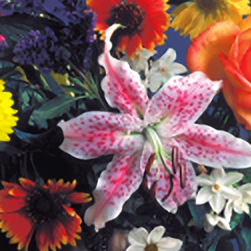}}
		\subfigure[RAISR]{\includegraphics[width=0.24\linewidth]{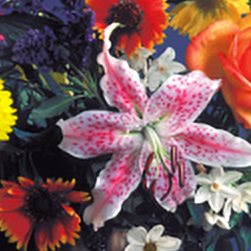}}
		\\
		\subfigure[Bicubic]{\includegraphics[width=0.24\linewidth]{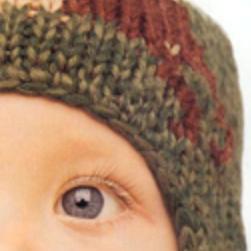}}
		\subfigure[SRCNN]{\includegraphics[width=0.24\linewidth]{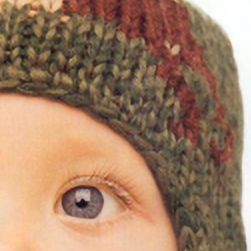}}				
		\subfigure[A+]{\includegraphics[width=0.24\linewidth]{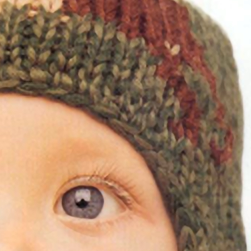}}
		\subfigure[RAISR]{\includegraphics[width=0.24\linewidth]{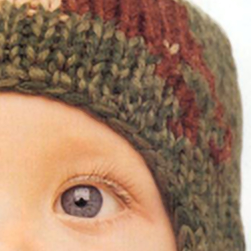}}
		
	\end{center}
	\vspace{-0.1in}
	\caption{Visual comparison for upscaling by a factor of 2 for the images \textsf{Zebra}, \textsf{Flowers} and \textsf{Baby}. (a,e,i) Bicubic interpolation, (b,f,j) SRCNN, (c,g,k) A+ with 1024 atoms, and (d,h,l) RAISR.}
	\label{fig:comp_vis}
\end{figure}

A quantitative comparison between RAISR and the leading SISR methods for $ 2\times $, $ 3\times $, and $ 4\times $ upscaling is given in Fig. \ref{fig:comp}. Per each method and upscaling factor we measure the average PSNR and SSIM over the images of Set5 and Set14 (the horizontal coordinate in Fig. \ref{fig:comp}) vs. the average runtime (the vertical coordinate in Fig. \ref{fig:comp}). The size of each point equals to $ 10\cdot\sigma_{\text{PSNR}}^{\text{err}}$ and $ 10^3\cdot\sigma_{\text{SSIM}}^{\text{err}} $, where $ \sigma^{\text{err}} $ is the standard error of the PSNR/SSIM. As can be inferred from Fig. \ref{fig:comp}, RAISR is competitive with the state-of-the-art methods in terms of these quality measures while being much faster. More specifically, for all upscaling factors, RAISR has similar restoration quality to the fast version of A+ (using 16 atoms), and it outperforms NE+LS, Zeyde et al., GR, and ANR. Notice that the more complex version of A+ (the one that uses 1024 atoms) and SRCNN performs better than RAISR, but with the cost of increased computations.

In terms of runtime, RAISR is the fastest method by far, demonstrating that high-quality results can be achieved without sacrificing the computational complexity. Following Fig. \ref{fig:comp}, our implementation is about one to two orders of magnitude 
faster than the baseline methods
(please refer to the \href{https://goo.gl/D0ETxG}{supplementary material} for detailed results). The runtime is evaluated on a 3.4GHz 6-Core Xeon desktop computer\footnote{Notably, we implemented a version of RAISR running on a mobile GPU, which runs at speeds of over 200Mpix/s. The desktop GPU version of the same can be expected to perform an order of magnitude faster.}.

\begin{figure}
	\begin{center}
		\subfigure[Original]{\includegraphics[trim={0.0cm 0.0cm 0.0cm 2cm},clip,width=0.49\linewidth]{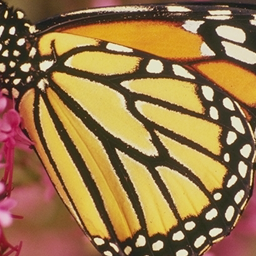}} \hfil
		\subfigure[Bicubic]{\includegraphics[trim={0.0cm 0.0cm 0.0cm 2cm},clip,width=0.49\linewidth]{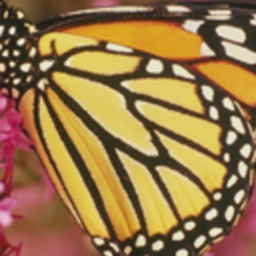}}
		\\
		\subfigure[Filtered image (learning without sharpening)]{\includegraphics[trim={0.0cm 0.0cm 0.0cm 2cm},clip,width=0.49\linewidth]{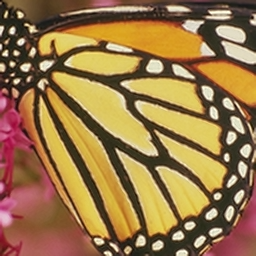}} \hfil
		\subfigure[Filtered image (learning with sharpening)]{\includegraphics[trim={0.0cm 0.0cm 0.0cm 2cm},clip,width=0.49\linewidth]{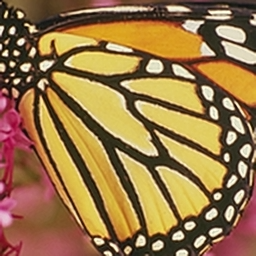}}
		\\
		\subfigure[Blending of (b) and (c)]{\includegraphics[trim={0.0cm 0.0cm 0.0cm 2cm},clip,width=0.49\linewidth]{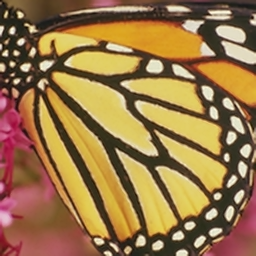}} \hfil
		\subfigure[Blending of (b) and (d)]{\includegraphics[trim={0.0cm 0.0cm 0.0cm 2cm},clip,width=0.49\linewidth]{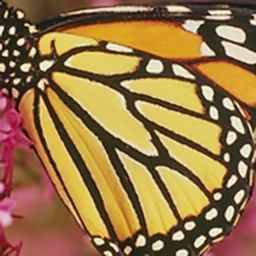}}
		
	\end{center}
	\vspace{-0.1in}
	\caption{{Visual illustration of the evolution of \textsf{Butterfly} image throughout the different blocks of RAISR, along with a demonstration of the built-in sharpening effect. (a) The original HR image, (b) Bicubic interpolation by a factor of 2 of the input LR image, (c) Filtered image -- the filters do not include a built-in sharpening, (d) Filtered image -- the filters map the initial interpolated image to its sharpened HR version, (e) blending result of the image in (b) with the one in (c), and (f) blending result of the image in (b) with the one in (d).}}
	\label{fig:intermediates}
\end{figure}


In the case of upscaling by a factor of 2, a visual comparison between RAISR, bicubic and the state-of-the-art A+ and SRCNN is provided in Fig. \ref{fig:comp_vis}.
As can be seen, the restoration quality of RAISR for both \textsf{Zebra}, \textsf{Flowers} and \textsf{Baby} images is competitive with much more complex algorithms. The ability of RAISR to restore continues edges is shown on the \textsf{Zebra} image. In addition, the effective reconstruction of fine details is demonstrated via the \textsf{Flowers} and \textsf{Baby} images.

{

\begin{table}[t]
	\centering
	\caption{{Quantifying the benefit of the built-in sharpening effect. We measure the average PSNR and SSIM (higher is better) over Set5 and Set14 images, along with the standard-error of each quality metric.}}
	\begin{tabular}{|c|c||c|c|c|c||c|c|c|c||c|c|} 
		
		\hline
		\multirow{2}{*}{\textbf{Dataset}} & \multirow{2}{*}{\textbf{Scaling}} & \multicolumn{4}{c||}{\textbf{Learning without sharpening}} & \multicolumn{4}{c||}{\textbf{Learning with sharpening}} & \multicolumn{2}{c|}{\textbf{Improvement}} \\
		\cline{3-12}          & & PSNR  & $ \sigma_{\text{PSNR}}^{\text{err}} $  & SSIM  & $ \sigma_{\text{SSIM}}^{\text{err}} $  & PSNR  & $ \sigma_{\text{PSNR}}^{\text{err}} $  & SSIM  & $ \sigma_{\text{SSIM}}^{\text{err}} $  & PSNR  & SSIM \\
		\hline
		\multirow{3}{*}{\textbf{Set5}} & 2$ \times $     & 35.913 & 1.374 & 0.947 & 0.015 & 36.153 & 1.337 & 0.951 & 0.015 & 0.241 & 0.004 \\
		\cline{2-12}          & 3$ \times $     & 32.061 & 1.392 & 0.895 & 0.019 & 32.211 & 1.363 & 0.901 & 0.019 & 0.150 & 0.006 \\
		\cline{2-12}          & 4$ \times $     & 29.689 & 1.479 & 0.839 & 0.021 & 29.837 & 1.481 & 0.848 & 0.021 & 0.147 & 0.009 \\
		\hline
		\multirow{3}{*}{\textbf{Set14}} & 2$ \times $     & 31.980 & 0.998 & 0.900 & 0.015 & 32.127 & 1.024 & 0.902 & 0.015 & 0.147 & 0.003 \\
		\cline{2-12}          & 3$ \times $     & 28.764 & 0.985 & 0.809 & 0.026 & 28.860 & 0.999 & 0.812 & 0.027 & 0.096 & 0.003 \\
		\cline{2-12}          & 4$ \times $     & 26.912 & 0.908 & 0.732 & 0.032 & 27.002 & 0.921 & 0.738 & 0.033 & 0.090 & 0.006 \\
		\hline
	
	\end{tabular}%
	\label{tab:w_wo_sharp}%
\end{table}%

Next, we test the effect of sharpening the HR training images. To this end, we learn a new set of filters using the same collection of 10,000 images, however this time without applying the pre-sharpening step. Following \mbox{Table \ref{tab:w_wo_sharp}}, this step indeed leads to an improvement both in PSNR and SSIM, emphasizing its potential. Motivated by these results, a promising future direction could be to test this concept as a way to improve various other state-of-the-art methods, e.g. A+ and SRCNN.

Fig. \ref{fig:intermediates} illustrates how the \textsf{Butterfly} image (taken from Set5) evolves throughout the different stages of RAISR, demonstrating the effect of the sharpening and the need for the blending step. The LR image is first upscaled by a factor of 2 using the bicubic interpolation (see Fig. \ref{fig:intermediates}b). Then, the pre-learned filters are applied: Fig. \ref{fig:intermediates}c shows the result when the filters are learned without sharpening, while Fig. \ref{fig:intermediates}d illustrates the case where the filters have a built-in sharpening effect. As can be seen, the image in Fig. \ref{fig:intermediates}d is indeed sharper than the one in Fig. \ref{fig:intermediates}c, and both have noticeable halos along the edges in addition to an amplification of the noise especially in flat areas. These artifacts are then reduced by the blending step, as depicted in Fig. \ref{fig:intermediates}e and \ref{fig:intermediates}f. In this case, the improvement achieved by the pre-sharpening during training is 0.44dB and 0.009 in terms of PSNR and SSIM, respectively.}

To conclude, we tested the upscaling performance of the proposed method on popular datasets and for various upscaling factors.
As demonstrated, RAISR is much faster than the leading algorithms, while achieving a competitive restoration performance.

\subsection{All in One Enhancement}
\label{sec:all_in_one}

The ability of the proposed algorithm to handle real-world scenarios is evaluated as well. More specifically, we test the ability of RAISR to upscale an image while reducing compression artifacts and improving the overall contrast of the image. The results are obtained using the scheme that is given in Fig. \ref{fig:RAISR_CT}. Differently from the conventional SISR problem (Section \ref{sec:sisr}), in this case we do not apply the back-projection step. In addition, in order to further reduce computations, we choose the bilinear as the initial interpolator. As a result, we manage to reduce the runtime by about a factor of 2, i.e, 
\emph{this version of RAISR is more than two orders of magnitude faster than the state-of-the-art methods A+ and SRCNN.}

In the learning stage, we downscaled the training images (taken from BSDS300 database \cite{martin2001database}) using the bicubic interpolation, followed by a compression step using JPEG, with a quality parameter that is set to 85 out of 100. In order to achieve contrast enhancement effect, the target HR images are pre-processed using the proposed CT-based DoG sharpener (see Fig. \ref{fig:CT_DoG}), where the blending is done by counting the modified bits of the CT descriptor, which allows the amplification of a wide range of frequencies. Naturally, the wider the Gaussian filters that are used in the DoG scheme, the wider the range of frequencies that are amplified. As such, we suggest using 2 DoG layers: The first layer applies a narrow Gaussian filter, with $ \sigma \approx 0.85 $ (i.e., the separable one dimensional  $ [1, \ 2, \  1]/4 $ filter), while the second is a much larger Gaussian filter of size $ 64\times64 $, with $ \sigma = 8.5 $ (i.e., we set $ \alpha=10 $). In both cases we use $ \rho = 55 $. Note that the hyper parameters of RAISR (filter-size, hashing parameters, etc.) are the same as in Section \ref{sec:sisr}.

\begin{figure}
	\begin{center}
		\subfigure[]{\includegraphics[width=0.32\linewidth]{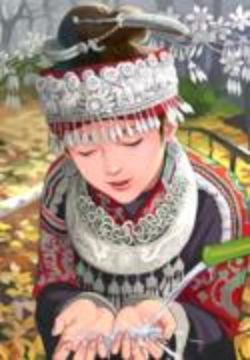}}
		\subfigure[]{\includegraphics[width=0.32\linewidth]{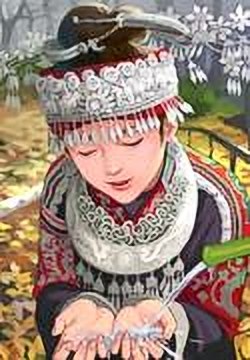}}
		\subfigure[]{\includegraphics[width=0.32\linewidth]{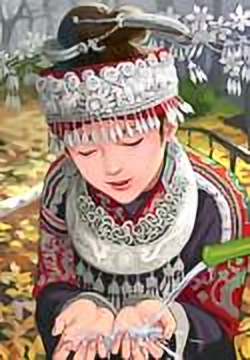}}
	\end{center}
	\vspace{-0.1in}
	\caption{Visual comparison for upscaling by a factor of 2 the JPEG-compressed \textsf{Comic} image (taken from Set14). (a) Bilinear interpolation, (b) RAISR with learning stage as in Section \ref{sec:sisr} which does not include compression of the training LR images, (c) RAISR with learning that involves compression of the training LR images.}
	\label{fig:comp_regular_to_all}
\end{figure}

\begin{figure}
	\begin{center}
		\subfigure[SRCNN]{\includegraphics[width=0.32\linewidth]{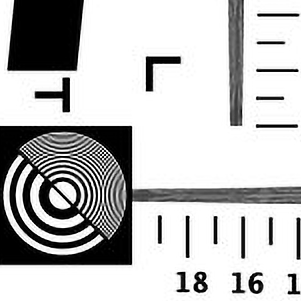}}
		\subfigure[A+]{\includegraphics[width=0.32\linewidth]{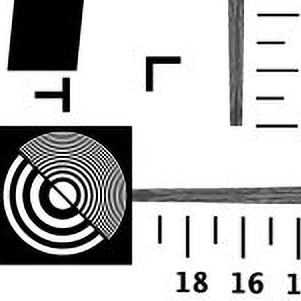}}
		\subfigure[RAISR]{\includegraphics[width=0.32\linewidth]{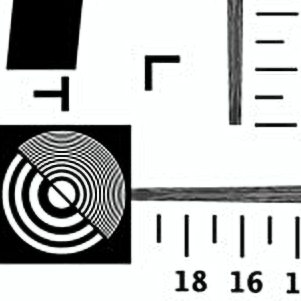}}
		\\
		\subfigure[SRCNN]{\includegraphics[width=0.32\linewidth]{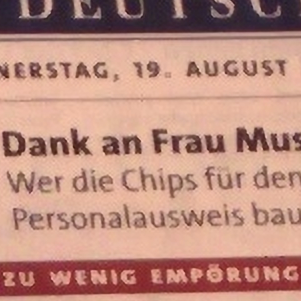}}				
		\subfigure[A+]{\includegraphics[width=0.32\linewidth]{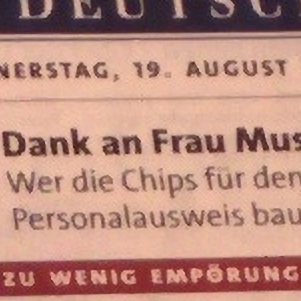}}
		\subfigure[RAISR]{\includegraphics[width=0.32\linewidth]{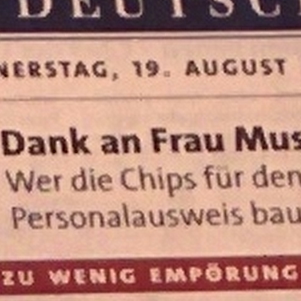}}
		\\
		\subfigure[SRCNN]{\includegraphics[width=0.32\linewidth]{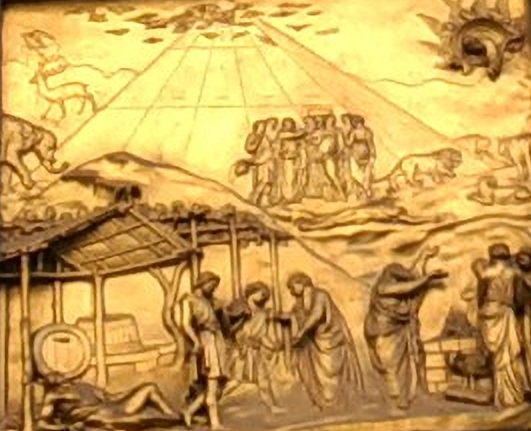}}				
		\subfigure[A+]{\includegraphics[width=0.32\linewidth]{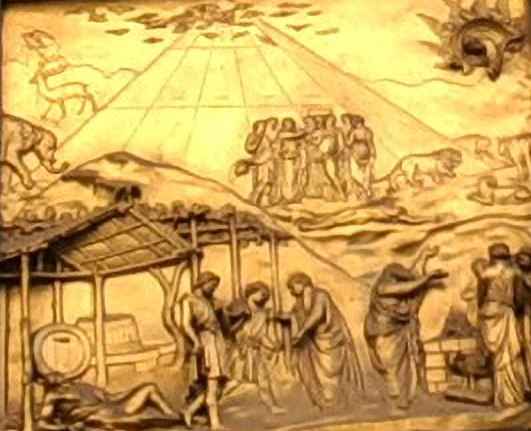}}
		\subfigure[RAISR]{\includegraphics[width=0.32\linewidth]{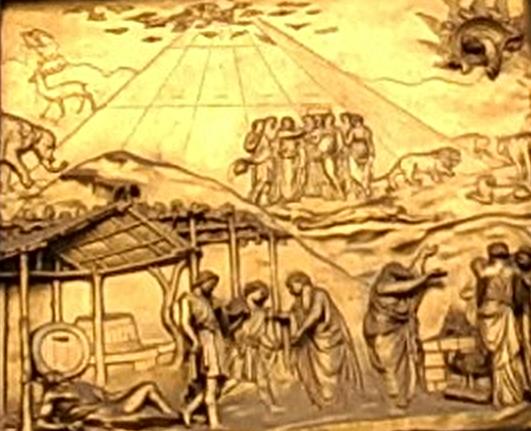}}
		
	\end{center}
	\vspace{-0.1in}
	\caption{Visual comparison for upscaling by a factor of 2 of \textsf{Resolution-Chart}, \textsf{Newspaper} and \textsf{Painting} images. (a,d,g) SRCNN, (b,e,h) A+ with 1024 atoms, and (c,f,i) RAISR. In this case, RAISR filters map compressed LR images to their contrast-enhanced HR versions.}
	\label{fig:comp_real}
\end{figure}

Before testing RAISR in a real world scenario, where the degradation model is unknown, we show the benefits of compressing the training LR images on a synthetic example. To this end, following Fig. \ref{fig:comp_regular_to_all}, we degrade the \textsf{Comic} image (taken from Set14) by downscaling it by a factor of 2 in each axis and then apply a JPEG compression on the result. As can be seen in Fig. \ref{fig:comp_regular_to_all}b, when the filters are learned without compressing the training LR images we obtain a sharp result but with the cost of undesired amplification of compression artifacts. On the other hand, when the learning stage includes compression, the upscaled outcome has less artifacts without the loss of sharpness, as illustrated in Fig. \ref{fig:comp_regular_to_all}c. This result should not surprise us as we take into account the degradation model that includes both downscaling (by a factor of 2) and compression. Quantitatively, compared to a learning stage that does not involve such a compression, in terms of PSNR we achieve an average improvement of 0.43dB and 0.34dB for Set5 and Set14, respectively. Similarly, we obtain higher SSIM score, where the average improvement is of 0.017 and 0.012 for Set5 and Set14, respectively.

Fig. \ref{fig:comp_real} shows the superiority of handling compression artifacts along with the visually pleasant outcome of our contrast-enhancement learning scheme. Upscaling results of a cropped and compressed version of the \textsf{Resolution-Chart} image is given in Fig. \ref{fig:comp_real}a-\ref{fig:comp_real}c, obtained by applying SRCNN, A+, and RAISR, respectively. As can be seen, RAISR reduces some of the compression artifacts (especially around the digits and in between the circles, located in the bottom-right and left parts of the image, respectively). Despite the reduction of compression artifacts, RAISR successfully keeps the desired sharpness property (notice the effective restoration of the very fine details of the circles in the left part of the image). 

The ability of RAISR to handle compressed, blurred with an unknown (possibly motion-blur) kernel, and noisy images is shown in \mbox{Fig. \ref{fig:comp_real}d-\ref{fig:comp_real}i}. As can be inferred, the blending step and the built-in suppression of compression artifacts increase the robustness of the algorithm, especially when tackling noisy images. Furthermore, the contrast-enhancement effect leads to better looking images (notice the contrast-enhanced letters in the \textsf{Newspaper} image and the amplification of the fine details in the \textsf{Painting} image). Additional examples on real images are shown in \href{https://goo.gl/D0ETxG}{supplementary material}.

To summarize, we demonstrated the ability of RAISR to achieve a high-quality restoration in various scenarios. Our experiments indicate that the proposed method stands in a line with the best algorithms that are currently available, while being about 2 orders of magnitude faster.

\section{Conclusion}
\label{sec:conclusion}

In this paper we proposed a rapid and accurate learning-based approach for single image super-resolution (called RAISR). The suggested algorithm requires a relatively small set of training images to produce a low-complexity mechanism for increasing the resolution of any arbitrary image not seen before. The core idea behind RAISR is to learn a mapping from LR images to their HR versions. The mapping is done by filtering a ''cheap'' upscaled version of the LR image with a set of filters, which are designed to minimize the Euclidean distance between the input and ground-truth images. 

More specifically, RAISR suggests a highly efficient locally adaptive filtering process, where the low-complexity is kept thanks to an appealing hashing scheme: In the learning stage, a database of images is divided into buckets (''cheap'' clusters) of patches that share similar geometry, and a filter is learned per each bucket. In the application side, based on the geometry of the LR patch (hash-table key), the relevant pre-learned filter (hash-table entry) is chosen and applied on the patch. 

Moreover, artifact-free results are obtained by blending the initial (cheap) estimation of the HR image with the filtered one. This step is based on the observation that in flat areas the reconstruction of the cheap upscaler is effective (there are no fine details or edges to be recovered in these areas). We harness the efficient CT descriptor for this task. As such, with a negligible computational cost, an accurate reconstruction is achieved. Note that two different blenders are suggested; the first allows the amplification of high frequencies only (using the ''randomness'' mask), while the second allows the enhancement of a wide range of frequencies. Differently from the randomness mask, the latter blending mechanism counts the number of bits of the CT descriptors that were changed due to the filtering step.

As such, the additional complexity of RAISR over a very basic interpolation method (e.g. bilinear interpolation) is roughly the application of 2-3 linear filters on the image: The first filter leads to the hash-table-key, pointing to the second filter that enhances the quality of the patch (the pre-learned filter), and the last filter (which is almost negligible) is used for the blending step. Despite the extremely low computational cost, the restoration performance of RAISR is competitive with much more complex state-of-the-art methods. For example, in the case of A+ (which is one the fastest methods that currently available), the complexity of choosing the filter is linear with the size of the dictionary, where  the complexity of our approach for choosing the filter is constant due to the hashing mechanism.

Furthermore, sharpening and suppression of compression artifacts are achieved by pre-processing the training images. A sharpening/contrast-enhancement effect is obtained by amplifying the details and contrast of the HR training images. Similarly, compression artifacts are handled by compressing the LR training images.
This results in filters that are unique in profile, and perform the dual purpose of de-aliasing and sharpening in a single filter. The pre-learned filters tend to have a smoothing effect in the center of the kernel to perform aliasing removal, while containing ridges near the edges of the kernel which have the effect of sharpening that edge. As such, our learning approach is able to learn an effective set of filters that would be nearly impossible to construct by hand.

Motivated by the sharpening effect that can be achieved, we suggested a novel sharpener, which is based on applying (separable) DoG filters on the image, which are capable of flexibly enhancing a wide range of frequencies. As a way to reduce artifacts (e.g. halos and noise amplification), we suggest using once again the CT blender, leading to an extremely efficient and effective sharpener.

We should note that the quality of the hashing mechanism is crucial. Therefore, a further study of the hash function is needed, and could possibly lead to improved results. In addition, we use a simple least squares solver to learn the filters, while improved results may be achieved by regularizing the learning with efficient priors. In a wider perspective, we wish to explore the ability of RAISR to cast as a boosting mechanism \cite{talebi2013saif, milanfar2013tour, romano2013improving, talebi2014global, kheradmand2014general, romano2015patch, romano2015boosting, ghimpeteanu2016decomposition}, where in this case the learned filters can be designed to map any input images (not only the ''cheap'' interpolated images) to the desired outputs.

\appendix

\subsection{Back Projection}
\label{app:back}
As a way to reduce artifacts and estimation errors, many methods (e.g., \cite{park2003super, dong2009nonlocal, song2010improved, yang2010image, yang2012coupled,he2013beta, makwana2013single}) harness the popular Iterative Back Projection (IBP) \cite{irani1991improving} as a global post processing step. Following the degradation model in Eq. \eqref{eq:degradation}, one can demand an equivalence between the LR image and the downscaled version of the estimated HR image. More formally, given an estimation of the HR image $ \tilde \x $ (i.e., the output of RAISR), we suggest minimizing the following cost
\begin{equation}
\label{eq:bp}
\min_{\x} \| \x - \tilde \x \|_2^2 \ \ \  \text{s.t.} \ \ \  \z = \D_{s}\H\x,
\end{equation}
leading to the desired HR estimation $ \xh $. The solution of Eq. \eqref{eq:bp} can be obtained by applying several iterations of gradient descent. We found it helpful to apply this post-processing step in the case of the conventional SISR process (the results in Section \ref{sec:sisr}).

\bibliographystyle{IEEEtran}   
\bibliography{refs}

\end{document}